%% file: root.tex
\documentclass{ieeetj}
\usepackage{cite}
\usepackage{amsmath,amssymb,amsfonts}
\usepackage{algorithm}
\usepackage{algpseudocode}
\usepackage{graphicx,color}
\usepackage{textcomp}
\usepackage{xcolor}
\usepackage{hyperref}
\let\labelindent\relaxt
\usepackage{enumitem}
\hypersetup{}

\def\BibTeX{{\rm B\kern-.05em{\sc i\kern-.025em b}\kern-.08em
    T\kern-.1667em\lower.7ex\hbox{E}\kern-.125emX}}
\AtBeginDocument{\definecolor{tmlcncolor}{cmyk}{0.93,0.59,0.15,0.02}\definecolor{NavyBlue}{RGB}{0,86,125}}

\def\seclogo{\vspace{10pt}}

\def\authorrefmark#1{\ensuremath{^{\textbf{#1}}}}

\begin{document}
% \receiveddate{XX Month, XXXX}
% \reviseddate{XX Month, XXXX}
% \accepteddate{XX Month, XXXX}
% \publisheddate{XX Month, XXXX}
% \currentdate{XX Month, XXXX}
% \doiinfo{XXXX.2022.1234567}

% \markboth{}{Author {et al.}}  
\title{Visuomotor Robotic Pruning in Planar Orchards Using Hybrid Reinforcement Learning}

\author{Abhinav Jain\authorrefmark{1}, Cindy Grimm\authorrefmark{1}, and Stefan Lee\authorrefmark{1}}
\affil{Collaborative Robotics and Intelligent Systems (CoRIS) Institute, Oregon State University, Corvallis OR 97331, USA}
\corresp{Corresponding author: Abhinav Jain (email: contact.abhinav.jain@gmail.com).}
\authornote{This research was supported by NSF and USDA-NIFA award no. 2021-67021-35344 (AgAID institute). }

\input{macros}

\input{sections/0_abstract}

\maketitle

\input{sections/1_introduction}

\input{sections/2_related_work}

\input{sections/3_0_simulator}

\input{sections/3_1_synthetic_trees}
\input{sections/3_2_robot_description}
\input{sections/3_3_setup}
\input{sections/3_4_defining_success}

\input{sections/4_0_setting_up_learning}
\input{sections/4_1_training_episodes}
\input{sections/4_2_state_action}
\input{sections/4_3_policy}

\input{sections/4_4_rewards}
\input{sections/4_5_synthetic_data}

\input{sections/5_0_learning_algorithm}
\input{sections/5_1_problem_definition}
\input{sections/5_2_ppo}
\input{sections/5_3_hybrid_ppo}

\input{sections/6_training}

\input{sections/7_deployment}

\input{sections/8_0_experiment_setup}
\input{sections/8_1_algorithm_validation}
\input{sections/8_2_policy_eval}
\input{sections/8_3_real_world_eval}
\input{sections/8_4_real_world_baseline}

\input{sections/9_0_experiments}
\input{sections/9_1_algorithm_experiments}
\input{sections/9_2_sim_experiments}
\input{sections/9_3_real_world_experiments}
\input{sections/9_4_real_world_baseline}

\input{sections/12_discussion}
\input{sections/13_conclusion}

\bibliographystyle{IEEEtran}
\bibliography{bibliography}
\begin{IEEEbiography}[{\includegraphics[width=1in,height=1.25in,clip,keepaspectratio]{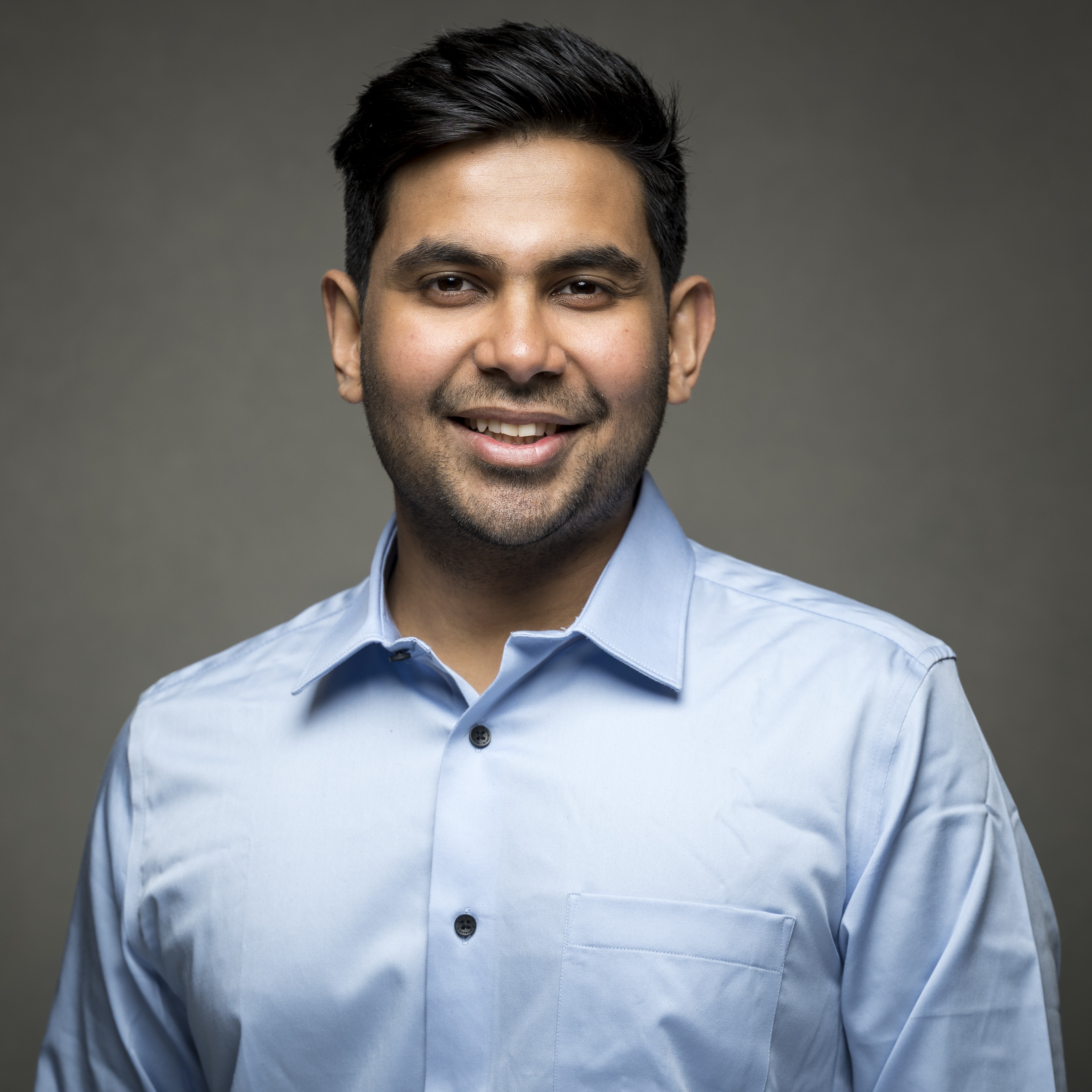}}]
{Abhinav Jain}~is a Ph.D. candidate and Graduate Research Assistant with the Collaborative Robotics and Intelligent Systems (CoRIS) Institute at Oregon State University, Corvallis, OR, USA. His research focuses on robotic manipulation in unstructured environments, with the goal of building dexterous embodied systems. His interests include real-to-sim high-fidelity simulations, reinforcement learning, and embodied AI. 

He received the B.Tech. degree in electronics and communication engineering from the Sardar Vallabhbhai National Institute of Technology (SVNIT), Surat, India, in 2020, and the M.S. degree in robotics from Oregon State University in 2026. He is currently pursuing the Ph.D. degree in robotics with a minor in artificial intelligence at Oregon State University. His work has been presented at AAAI, ICRA, and workshop at CVPR.
\end{IEEEbiography}

\begin{IEEEbiographynophoto}
{Cindy Grimm,} works in the area of robotic grasping and manipulation for agriculture, as well as ethics, law and policy related to robotics. Focus areas are tree fruit picking and pruning. Her previous projects include: modeling the developing heart, understanding how the shape of bat ears influences their sonar patterns, art-based rendering, 3D sketching, and interfaces for 3D medical image segmentation. She received her PhD from Brown University in 1995 in the area of surface modeling, spent two years working at Microsoft Research on facial animation, then ten years as faculty in Computer Science at Washington University in St. Louis. She is now in the Mechanical Engineering department at Oregon State University.
\end{IEEEbiographynophoto}

\begin{IEEEbiographynophoto}
{Stefan Lee,} is an Associate Professor in the School of Electrical Engineering and Computer Science at Oregon State University, where he holds the Brent and Elaine Leback Professorship in Engineering and co-directs the Dynamic Robotics and AI Lab. His research focuses on embodied AI: building agents that can see, talk, and act, drawing on computer vision, natural language processing, and deep learning. He received his Ph.D. in Computer Science from Indiana University in 2016 and was a research scientist at Georgia Tech before joining Oregon State in 2019. His honors include an NSF CAREER Award (2024), the Engelbrecht Early Career Award, an ICLR 2023 Outstanding Paper Award, and a Best Paper Award at EMNLP 2017.
\end{IEEEbiographynophoto}

\vfill\pagebreak

\end{document}

%% file: macros.tex
% Custom macros for the paper
\newcommand{\aj}[1]{\textcolor{red}{#1}}
% Perpendicular projection operator
\newcommand{\perpproj}[2]{\text{proj}_{\perp #2}(#1)}

%% file: sections/0_abstract.tex
\begin{abstract}
Dormant tree pruning is labor‑intensive yet essential for maintaining modern high‑productivity fruit orchards. In this work, we focus on pruning of modern planar tree training systems---V-Trellis apples and UFO cherries---where trunks and primary branches are trained into approximately planar walls. We introduce an end-to-end pipeline to learn a closed-loop visuomotor controller for robotic pruning. This controller is trained entirely using simulation and synthetically generated data and deployed in real orchards in a zero-shot manner. The pipeline comprises synthetic generation of planar orchard tree meshes, construction of a physics‑based orchard simulator, automated collection of successful pruning trajectories via motion planning, and policy learning with a novel hybrid reinforcement‑learning algorithm that combines offline demonstrations with online simulated rollouts. The controller uses optical-flow inputs from a wrist-mounted camera---avoiding the need for full 3D-reconstruction---and continuously guides the cutter through cluttered branch environments to a specified cutpoint with correct tool orientation. 
In exhaustive simulated task-space evaluations over 3,000 pruning points, the policy attains 49.9\% success on V-Trellis apples and 46.0\% on UFO cherries. We validate the learned controller across 38 physical trials—comprising 28 outdoor field trials in commercial and experimental orchards and 10 indoor laboratory tests—demonstrating zero-shot sim-to-real transfer. The learned policy also outperforms a classical RRT-Connect baseline on physical hardware in laboratory trials.
% Finally, we evaluate the proposed Hybrid-PPO algorithm against PPO and PPO with behavioral cloning, observing improvements in mean success rates.
\end{abstract}


%% file: sections/1_introduction.tex
\section{INTRODUCTION}
\label{sec:introduction}
\IEEEPARstart{M}{odern} farming techniques have adopted carefully designed tree structures that improve productivity and labor efficiency, but must be maintained through cycles of tree pruning and tying trees to specific shapes during the dormant winter season. In this study, we focus on robotic pruning of two such structures: (i) Envy apple trees in a \emph{V-Trellis} structure and (ii) cherries in an \emph{Upright Fruiting Offshoot} (UFO) structure. These structures are shown in Figure~\ref{fig:cluttered_env}. Tree trunks are grown in  rows and consist of primary branches tied horizontally (V-Trellis) or vertically (UFO) to a set of fixed wires running parallel to one another and supported by posts. This gives rise to an approximately planar shape for each tree. These primary branches grow spurs -- the sites of fruit production -- and tertiary branches, both of which require pruning.
\input{figures/1_robot}

\input{figures/2_real_env}
Dormant tree pruning is labor-intensive, costing up to 25\% of annual labor costs in high-density apple orchards~\cite{pruningcost,wsuPruning,infaco2022farm}. These costs are projected to rise due to declining immigration of farm laborers to the United States, who have historically fulfilled this agricultural labor demand~\cite{nae2019immigrants,agamerica2022labor,daniels2018strawberries}.
Additionally, hand pruning leads to inconsistency, which affects fruit quality and overall yield~\cite{bates_overpruning}.
	
% value proposition if problem is solved
Robotic pruning has the potential to address this labor shortage and produce consistent and reproducible pruning outcomes. However, there are numerous challenges in building a holistic robotic system to perform pruning, including: i) perceiving the tree structure, ii) determining which branches to cut and where to cut them, and iii) driving the pruner attached to the branch-cutting robot to the desired cut location {\em without} colliding with rigid tree structures or supports. This paper focuses on the third challenge: learning a vision-based control policy to guide a robotic pruner to reach the given cutpoint. Our robot consists of a UR5e~\cite{ur5e} arm equipped with a pruning tool and an eye-in-hand camera mounted at the end-effector. The arm is mounted on a Farm-NG Amiga mobile base~\cite{farmng_amiga}. The complete robot is shown in Figure~\ref{fig:robot} (left) and the goal is to drive the end-effector to the target pruning point as illustrated in Figure~\ref{fig:robot} (right).

% why no one solved it yet
% Robotic pruning in modern fruiting tree orchards (e.g., apple orchards) presents a challenging setting for robotic control.
Vision based control is difficult in an orchard as individual tree geometries are unique and can be intricate --- introducing high variation. In addition, the presence of other structures such as trellis wires and posts contributes to significant visual and physical clutter. Figure~\ref{fig:cluttered_env} shows the `V' structure in which both V-Trellis and UFO architectures are grown, along with the clutter due to various orchard elements. Classical motion planning techniques rely on accurate 3D reconstruction of the tree to perform the pruning task~\cite{CorbettDaviesStraddleVines,trimbot,SilwalBumblebee}. However, generating these 3D models is non-trivial: LiDAR-based solutions can be prohibitively expensive, whereas cheaper depth sensing technologies that rely on structured-light or coded-light (e.g. Intel RealSense D435) struggle in outdoor environments due to interference from sunlight and surface scattering~\cite{lidaroutdoorbad}. Moreover, thin branches and a dynamic outdoor environment further complicate accurate reconstruction. Beyond perception, trees are large, and pruning cuts must be made at particular angles relative to the branch. This increases the required workspace size and demands a wider range of target end-effector orientations along with obstacle avoidance than standard reaching tasks. At the same time, effective solutions must operate quickly to make an impact -- human pruners average one cut per second and perform 10-50 cuts per tree~\cite{FlynnPruningDecision}. Thus, computationally intensive 3D reconstruction pipelines or planning techniques may result in solutions that are inaccurate or slow for practical adoption. 

% what we do to solve it
To avoid this explicit 3D reconstruction, Jain et al.~\cite{JainRL} proposed a reinforcement learning–based visuomotor policy that enabled a 6-DOF robotic arm equipped with a cutter to reach designated pruning points using only wrist-mounted camera images. Their approach avoided collisions and oriented the cutter perpendicular to the target branch without requiring explicit 3D reconstruction, achieving a 30\% success rate and establishing a foundation for zero-shot sim-to-real transfer of visuomotor pruning policies. Specifically, they developed a formal-grammar–based pipeline for generating realistic tree geometries, constructed a sufficiently realistic orchard simulation environment for training, designed a task-specific reinforcement learning reward function, and adopted optical flow instead of RGB images for perception to enable transferable learning without requiring photorealism~\cite{OpticalFlowAlex2022IROS}.

However, the performance of this controller was limited to branches oriented approximately vertically (up/down) due to the robot’s kinematic constraints and exploration bottlenecks inherent to reinforcement learning. Moreover, the controller was never evaluated in real orchards. In this work, we address these limitations through five key extensions: (i)  we perform a reachability study to reposition the pruner to improve our working space; (ii) we develop an automated motion-planning pipeline that generates successful pruning trajectories to be used as demonstrations; (iii) we introduce a novel hybrid reinforcement learning algorithm based on PPO~\cite{Schulman2017ProximalPO} that incorporates synthetically generated demonstrations to improve training efficiency and policy performance; (iv) we train and evaluate the policy across two tree architectures---V-Trellis and UFO---to demonstrate adaptability of our method; and (v) we validate the approach in real-world orchard trials in the USA, showing sim-to-real transfer in actual orchards. These additions improve the performance of the learned policy and make up a complete sim-to-real pipeline for vision-based robotic pruning, moving us closer to real-world deployable systems.

% how well does our solution work

We evaluate our controller's performance both in simulation and physical environments. In simulation, we conduct an exhaustive task-space evaluation across 3,000 cutpoints (placing targets at three distinct locations on 1,000 uniformly sampled branch orientations) using 100 trees unseen during training. On these test sets, our learned policy achieves a success rate of 49.9\% on V-Trellis apples and 46.0\% on UFO cherries. To compare, we run an oracle RRT-Connect planner with access to perfect state information and geometry reporting the maximum possible success rate as 93\% for V-Trellis and 95\% for UFO on the same test set. 

For physical evaluations to show sim-to-real transfer, we validate our policy across 38 total trials in two distinct settings: (i) 10 controlled indoor laboratory trials on a constructed V-Trellis tree model, and (ii) 28 outdoor field trials across commercial V-Trellis apple and experimental UFO cherry orchards under actual operational conditions. These cutpoints were selected within the reachable workspace and biased toward safer configurations that matched our training assumptions. In these outdoor trials, the policy demonstrates strong sim-to-real transfer, achieving 71\% success on V-Trellis apples (10/14) and 35\% on the experimental UFO architecture (5/14, despite task-space assumption violations caused by the steeper canopy structure). On the indoor laboratory setup, where the canopy geometry closely matched our training assumptions, the policy achieved a 90\% success rate (9/10).

Separately, to benchmark against classical motion planning on physical hardware, we conducted a direct comparison on a \emph{different} laboratory cutpoint set of 15 targets sampled for diversity in location and orientation (rather than biased toward safe, clearly reachable configurations). On this shared baseline set our policy achieved 46.7\% success (7/15), outperforming point-cloud-based RRT-Connect at 26.6\% (4/15), which failed primarily due to perception errors in the reconstructed mesh used for planning. These experiments demonstrate that our solution transfers zero-shot to real-world field conditions and outperforms the classical RRT-Connect baseline.

The remainder of this paper is organized as follows. Section~\ref{sec:related} reviews related work. Section~\ref{sec:meth:sim} details the simulation environment—including synthetic tree generation, task definition, and robot setup—while Section~\ref{sec:meth:setting_up_learning} outlines the learning framework, observation and action spaces, network architecture, reward formulation, and synthetic data generation. Section~\ref{sec:algorithm} introduces our proposed Hybrid-PPO algorithm, followed by details on policy training (Section~\ref{sec:training}) and physical deployment (Section~\ref{sec:deployment}). In Section~\ref{sec:experiments}, we define experiments to: (a) validate the Hybrid-PPO algorithm (Section~\ref{sec:experiments}.\ref{sec:experiments:algo}), (b) evaluate overall policy performance in simulation (Section~\ref{sec:experiments}.\ref{sec:experiments:policy}), (c) demonstrate sim-to-real transfer in orchards (Section~\ref{sec:experiments}.\ref{sec:experiments:real_world}), and (d) benchmark our policy against a classical RRT-Connect baseline (Section~\ref{sec:experiments}.\ref{sec:experiments:real_world_baseline}). Results are presented in Section~\ref{sec:results}. Finally, Sections~\ref{sec:limitations} and~\ref{sec:conclusion} discuss key limitations and conclusions.

%% file: figures/1_robot.tex
\begin{figure}[t]
    \centering
    \includegraphics[width=\columnwidth]{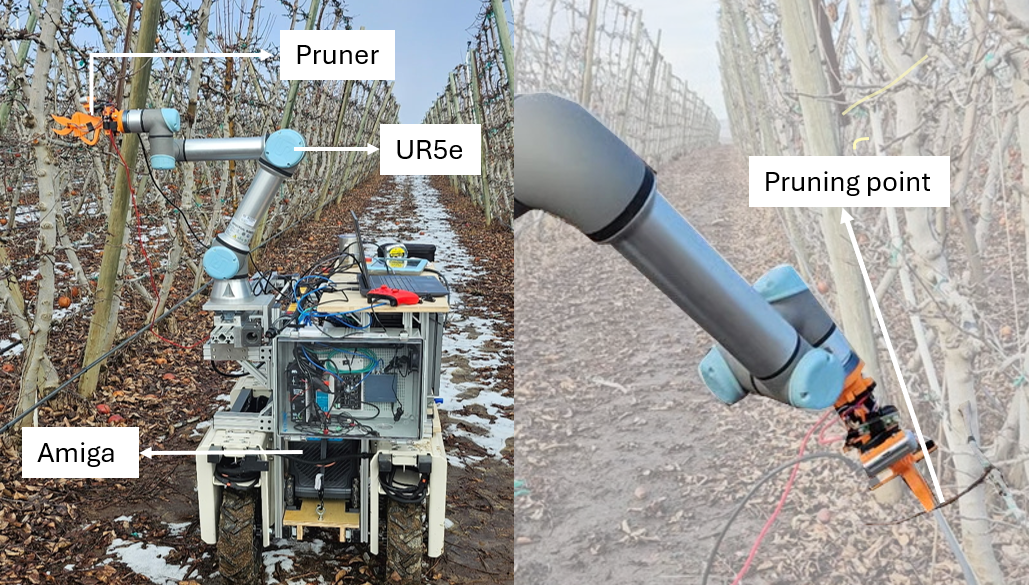}
    \caption{\textit{(Left)} Our pruning platform: a UR5e arm mounted on a Farm‑NG Amiga base with a custom pruning tool as the end‑effector. \textit{(Right)} Example final configuration from a pruning trial, with the target branch positioned at the bottom of the cutter jaws.}
    \label{fig:robot}
\end{figure}

%% file: figures/2_real_env.tex
\begin{figure*}[t]
    \centering
    \includegraphics[width=\textwidth]{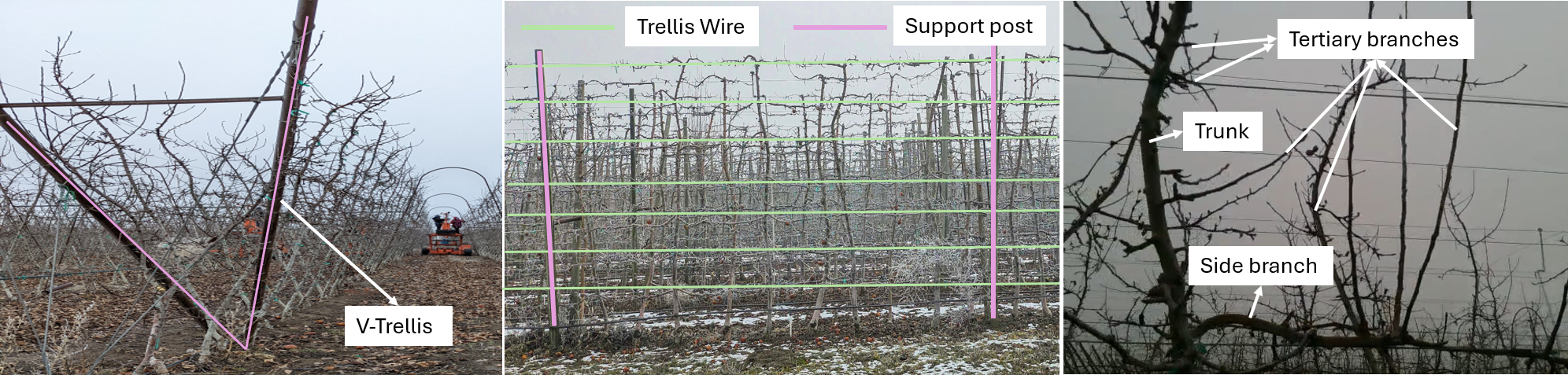}
    \caption{\textit{(Left)} The ``V'' shape of trees in V-Trellis and UFO architectures. 
            \textit{(Middle)} Orchard structures such as trellis wires (green) and support posts (pink). 
            \textit{(Right)} Different types of tree branches that contribute to visual and physical clutter.}
    \label{fig:cluttered_env}
\end{figure*}

%% file: sections/2_related_work.tex
\section{RELATED WORK}
\label{sec:related}
\vspace{4pt}\noindent\textbf{Robotic Pruning. }
Prior work in robotic pruning has focused on grapevine pruning using a traditional multi-step process. This involves: (i) constructing a point cloud from multiple multi-view depth images, (ii) determining which canes should be cut or trimmed, and (iii) employing a motion planner --- either RRT-Connect~\cite{CorbettDaviesStraddleVines,trimbot,SilwalBumblebee} or reinforcement learning (RL) policies~\cite{YandunRLPruning} to guide the cutter to the designated cutpoints.

To construct the point clouds, depth information is obtained from cameras that utilize structured or coded lighting or time-of-flight sensors. However, these sensors see a degradation in performance when used in outdoor environments due to strong sunlight or surface scattering~\cite{lidaroutdoorbad}. As a result, controlling lighting is crucial for generating accurate point clouds. For example, Corbett-Davies et al. \cite{CorbettDaviesStraddleVines} straddle the grapevine rows within a box-like structure, creating a controlled environment with custom lighting. In contrast, Silwal et al. \cite{SilwalBumblebee} eliminate the need for such physical enclosures by using an advanced lighting system to regulate image exposure and obtain consistent depth.

Motion planning to reach pruning points is generally achieved using sampling-based planners, such as RRT-Connect, which treat the point cloud as the obstacle map. Due to the large number of collision objects in orchard environments, running these planners can be computationally intensive. To address this, Silwal et al.~\cite{SilwalBumblebee} reduce planning time by relaxing the collision constraints and allowing collisions with non-rigid parts of the tree during motion planning. In contrast, Yandun et al.~\cite{YandunRLPruning} replace sampling-based planning altogether with reinforcement learning, developing a policy that directly maps robot proprioception and occupancy grid data to motor torques.

Transitioning from these methods used for grapevines to tree pruning presents several challenges. First, the geometry of trees is more difficult to capture than that of grapevines due to the range of scales. Tree trunks range from 10~cm to 30~cm in diameter, with tertiary branches as thin as 1~cm.  Second, these tertiary branches ``fill'' the space between support branches and are oriented in all directions, resulting in a large number of potential collisions. 
%As we find in our real-world experiments with point-cloud reconstruction (see Section \ref{sec:real_world_experiments}), these varying scales, combined with the undesired illuminations in outdoor settings, often result in inaccurate 3D reconstructions and failure of downstream planning.
Given these limitations, You et al.~\cite{AlexUFOPruning} proposed a novel approach for UFO tree pruning that bypassed point clouds altogether. Instead, they use a learning-based visual servoing policy to guide the end-effector to the pruning point. This policy minimizes the image-based distance between the pruning point and the cutter's mouth; however, it assumes that the branch is always perpendicular to the cutter and does not avoid collisions. Jain et al.~\cite{JainRL} extended the learning-based visual servoing approach to include 6DOF control along with obstacle avoidance. Nevertheless, the resulting policy’s performance remains below levels required for commercial deployment.

\vspace{4pt}\noindent\textbf{Hybrid RL.}
We use the term \emph{Hybrid RL} to refer to reinforcement learning approaches that integrate both online rollouts and a static set of offline collected demonstrations. The most direct way to use demonstrations is through Behavior Cloning (BC)~\cite{bc}, where the policy is trained by supervised learning on expert state–action pairs. While simple and effective with large, clean datasets, BC typically struggles in settings where demonstrations are limited, noisy, or low-coverage, often causing the learned policy to overfit to the demonstration distribution and degrade under distribution shift.

A common strategy to mitigate these issues is to combine BC with online RL by adding a behavior-cloning loss during policy optimization. This regularizes the policy toward expert behavior while still permitting reward-driven improvement. On-policy methods such as DAPG~\cite{rajeswaran2018dapg}, PPO+BC~\cite{ppo_bc}, and BC-SAC~\cite{lu2023imitation} augment the policy-gradient update with a BC term, placing the policy near expert behavior early in training and gradually reducing the influence of this term over time. PIRLNav~\cite{ramrakhya2023pirlnav} follows a related paradigm: it first trains the actor via supervised imitation, pre-trains the critic with on-policy rollouts, and then switches to standard RL without further reliance on the offline data.

However, when demonstrations are sub-optimal or inconsistent, mixing BC and RL losses can lead to conflicting gradients because the two objectives may push the policy in different directions. This issue is addressed by~\cite{qfilter}, which introduces a Q-filter mechanism that applies the BC loss selectively—only when the critic estimates that the demonstrator’s action is better than the current policy’s action. Moreover, RL algorithms depend on different learning signals at different stages of training, and abruptly reducing or removing the BC term can prevent the agent from fully benefiting from the available demonstrations.

A second class of methods addresses these limitations by incorporating demonstrations directly into an off-policy RL framework. Instead of adding BC losses to an on-policy update, these approaches store expert trajectories in a replay buffer and train value functions from a mixture of expert and online transitions. Value-based methods such as DQfD~\cite{hester2018deep} and HyQ~\cite{song2023hybridrl} bootstrap Q-functions from expert data using modified Bellman backups and margin-based or hybrid objectives. Actor–critic methods like DDPGfD~\cite{vecerik2017leveraging} and AWAC~\cite{nair2020awac} extend this idea by training the critic off-policy using both expert and self-generated transitions, while updating the actor through TD or advantage-weighted gradients. By repeatedly bootstrapping from expert transitions, these methods effectively leverage demonstration data during both pretraining and online fine-tuning.

While many offline RL methods have been adapted to incorporate demonstrations, PPO~\cite{Schulman2017ProximalPO}—a simple, stable, and widely used on-policy algorithm—is often overlooked in this context. Behavior Proximal Policy Optimization (BPPO)~\cite{zhuang2023bppo} shows that PPO’s clipped surrogate update, when applied with modest off-policy adjustments, can achieve competitive performance on offline RL benchmarks. Motivated by this, we propose \emph{Hybrid-PPO}, an extension of PPO that integrates expert demonstrations into the training process, aiming to use PPO’s stability and simplicity while using offline data.

%% file: sections/3_0_simulator.tex
\section{SIMULATING THE ROBOTIC TREE PRUNING TASK}
\label{sec:meth:sim}

To support visuomotor policy learning, we construct a simulator for robotic tree pruning that mimics the geometry of an orchard scene. As our policy relies on optical flow rather than image texture, photorealism is not required. Our goal is to generate diverse pruning scenarios with variation in cutpoint locations, branch orientations, and lighting, while maintaining structural similarity to modern orchard training systems. 
\input{figures/3_branch_curve}

%% file: figures/3_branch_curve.tex
\begin{figure}[t]
    \centering
    \includegraphics[width=\columnwidth]{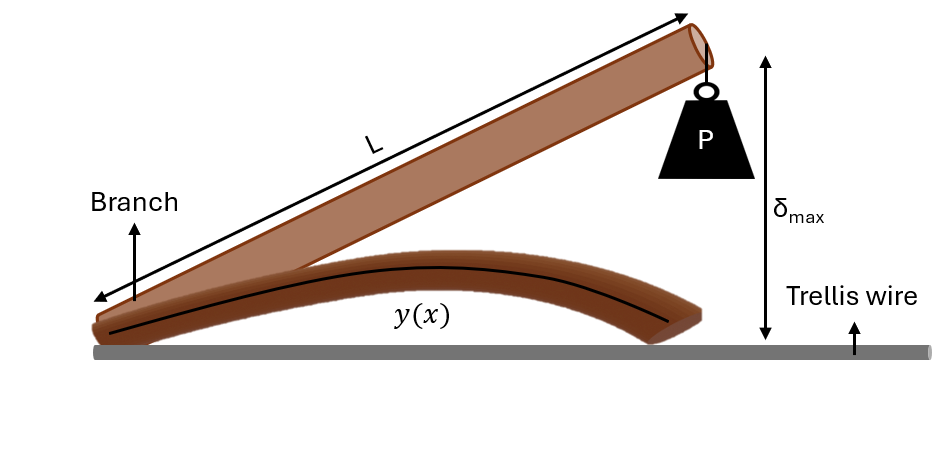}
    \caption{Modeling the branch as a cantilever beam with a fixed end for the tying process. A transverse load ($P_y$) is applied at the free end to obtain the resulting curvature $y_x$.}
    \label{fig:branch_curve}
\end{figure}

%% file: sections/3_1_synthetic_trees.tex
\subsection{Generating Synthetic Trees.} 
\label{sec:meth:sim:synthetic_trees}

\input{figures/4_synthetic_real_trees}

There are several common modern planar tree orchard structures --- e.g., V-Trellis, Upright Fruiting Offshoot, and Tall spindle~\cite{FlynnPruningDecision,MattUFO} --- that share the same characteristic structure. There are taut wires tied between posts to which trunks and/or branches are tied to induce specific shapes. This shaping is performed through annual cycles of tree growth, followed by pruning and branch tying. While there are several existing packages for ``growing'' natural trees programmatically, none allow for the simulation of this shaping process required to generate realistic orchards. Following Jain et al.~\cite{JainRL}, we extend an open-source plant growth software package (L-Py~\cite{lsystemsplants}), modifying its process to include cyclic pruning and tie-down phases during tree growth, referred to as \emph{Tree-L-Py} from here on. Along with tree meshes, Tree-L-Py also generates metadata that maps each face in the mesh to a tree-part label (Trunk, Primary Branch, Secondary Branch) for use in downstream tasks. Figure~\ref{fig:trees_real_synthetic_comparison} shows both the Tree-L-Py generated meshes and real-world UFO and V-Trellis trees. \emph{The tree generation code can be found here: https://github.com/OSUrobotics/lpy\_treesim.}

\vspace{4pt}\noindent\textbf{Tree architecture description.} We focus on the V-Trellis and UFO architecture; however, the Tree-L-Py software package can generate trees in any planar architecture by defining the correct grammar.

\textit{V-Trellis} often used for apples is characterized by alternate trunks tilted at 15 degrees to form a V-shape. A vertical trunk supports branches tied perpendicularly to wires spaced $\approx45$~cm apart; each wire carries one branch extending about 61~cm. These branches grow tertiary branches that are the fruiting sites for the tree. 

\textit{UFO} architecture is used for cherries and is grown on a `V' similar to V-Trellis but are trained with a horizontal trunk, from which upright secondary offshoots emerge approximately every 15~cm; these vertical shoots grow spurs, which are the fruiting sites. In both systems, pruning requires the removal of excess tertiary branches and fruiting sites to maintain productivity and tree structure. Figure \ref{fig:cluttered_env} (left) illustrates this `V' structure.

\vspace{4pt}\noindent\textbf{Pruning and tying rules.} To generate trees in simulation, we implement pruning and tie-down strategies specific to each system. For V-Trellis, we assign two branches per unoccupied horizontal wire, one on each side. For UFO, each horizontal wire supports a single upright offshoot. Side branches not tied in the previous growth cycle are removed, and new branches are tied down at pre-defined intervals.

\vspace{4pt}\noindent\textbf{Defining the curvature.} Branch curvature in Tree-L-Py from tie-down is approximated using classical beam theory. Each branch is modeled as a cantilever beam fixed at the base and subjected to forces at the free end. These forces flex the branch tip toward the tie-down point on the wire, and the resulting deflection profile defines the branch curvature in simulation. Figure~\ref{fig:branch_curve} illustrates this flexing process.

We define a \emph{vector-valued deflection} along the branch:
\[
\mathbf{y}(x) =
\begin{bmatrix}
y_x(x) \\
y_y(x) \\
y_z(x)
\end{bmatrix}, \quad x \in [0,L],
\]
where the branch deflection is computed independently along each axis. For each axis $i \in \{x, y, z\}$, the deflection $y_i(x)$ at a location $x$ is given by the standard cantilever beam equation:
\begin{equation}
y_i(x) = \frac{P_i x^2}{6 E I_i} (3 L - x)
\label{eq: bending}
\end{equation}
where $P_i$ is the end-point force along axis $i$, $I_i$ is the second moment of area about the bending axis, and $E$ is the modulus of elasticity. 

The distance $\delta_{\max,i}$ (Figure~\ref{fig:branch_curve}) is the desired deflection of the branch tip along axis $i \in \{x, y, z\}$, measured from the undeformed branch tip to the wire. This defines a boundary condition such that at a distance $L$ along the branch, the deflection satisfies $y_i(L) = \delta_{\max,i}$. The force needed to achieve this deflection along axis $i$ can then be computed by:

\begin{equation}
   P_i = \frac{3\delta_{\text{max,i}}EI_i}{L^3}
   \label{eq: bending_force}
\end{equation}

Using equation~\ref{eq: bending} and equation~\ref{eq: bending_force}, the deflection along each axis at any point of the branch can be calculated by:

\begin{equation}
   y_i(x) = \frac{\delta_{\max,i}}{2 L^3} \, x^2 (3 L - x), \quad i \in \{x, y, z\}
   \label{eq: deflection}
\end{equation}

Equation~\ref{eq: deflection} is independent of Young's modulus $(E)$ and second moment of area $(I)$, hence, no physical measurements from actual trees are required. This is extended over all three spatial dimensions to define the curve that a branch takes when tied down.

\vspace{4pt}\noindent\textbf{Mesh metadata.} We assign colors to the mesh triangles based on the tree part they represent, with labels corresponding to the trunk, primary branches, and secodnary branches. These semantic labels are later used to detect collisions with specific tree components and to identify tertiary branches as possible pruning candidates.

This framework allows us to generate an arbitrary number of trees with stochastic variation and labels and import them into the simulation environment. The tree geometry has not been formally verified to be statistically consistent with real orchards, but we have compared the resulting geometry to our extensive scans of orchard trees and confirmed our modeling choices with a horticultural expert. 

% Figure~\ref{fig:trees_real_synthetic_comparison} shows both the Tree-L-Py generated meshes and real-world UFO and V-Trellis trees.

%% file: figures/4_synthetic_real_trees.tex
\begin{figure*}[t]
    \centering
    \includegraphics[width=\textwidth]{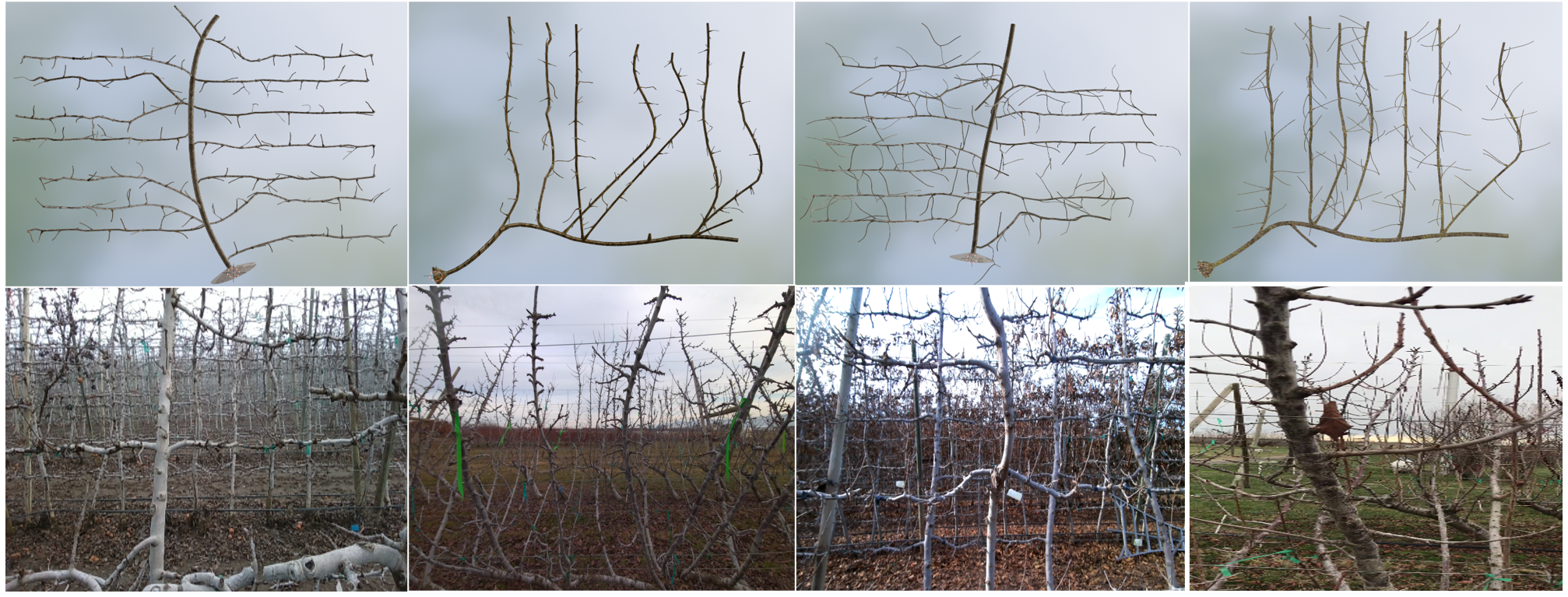}
    \caption{\textit{(Top row)} Tree-L-Py generated Envy apple and UFO trees. 
            \textit{(Bottom row)} Images of trees from a real orchard.}
    \label{fig:trees_real_synthetic_comparison}
\end{figure*}

%% file: sections/3_2_robot_description.tex
\subsection{Robot Description} 
\label{sec:meth:sim:robot}
Our pruning robot consists of a 6-DOF UR5e~\cite{ur5e} arm mounted on a Farm-NG Amiga~\cite{farmng_amiga} wheeled platform. We attach a custom pruning tool to the UR5e end-effector, consisting of a cutter with an integrated camera in an ``eye-in-hand'' configuration, as well as two VL53L4CD time-of-flight (ToF) range sensors mounted on either side. This tool is shown in Figure~\ref{fig:pruner_and_env}, while the complete robot platform is shown in Figure~\ref{fig:robot}. During operation, the robot is positioned approximately orthogonal to the tree.

Tool placement significantly influenced the arm’s reachable workspace. To determine the optimal tool position, we performed a reachability analysis within the defined workspace. Reachability is defined as the percentage of uniformly sampled pruning cutpoints (see sampling procedure in Sec.~\ref{sec:meth:sim}.\ref{sec:meth:setting_up_learning:sampling}) that are reachable by the robot in simulation. Using trees generated with Tree-L-Py, we evaluated reachability by running the sampling-based motion planner RRT-Connect to generate collision-free motion plans to all sampled cutpoints. We compared two mounting configurations: (a) an off-center configuration translated along the left–right axis, as used in \cite{JainRL}, and (b) a wrist-aligned tool configuration. The wrist-aligned configuration achieved substantially higher reachability (93\%) compared to the off-center configuration (65\%). Consequently, all experiments use the wrist-aligned cutter configuration.

%% file: sections/3_3_setup.tex
\subsection{Simulation setup}
\label{sec:meth:sim:sim_setup} We use the PyBullet~\cite{pybullet} physics simulator to create the orchard environment with the robot (UR5e on Amiga), Tree-L-Py generated trees, and other orchard structures such as trellis wires and posts. A virtual camera is also placed at the appropriate place on the cutter. Each element in the environment is assigned a unique texture. Additionally, more trees and a textured image of an orchard is placed behind the target tree to simulate the cluttered background of an orchard. Though not directly observed by the model, these textures are required to produce realistic optical flow images. The simulation scene can be seen in Figure~\ref{fig:pruner_and_env} (right). At the start of the simulation, noise is added to the robot orientation ($\pm 5$ degrees over its yaw, pitch, and roll axes) and camera placement ($\pm 2$ degrees pan and tilt) to mimic real-world conditions when setting up a robot in the field. This domain randomization technique makes the learned policy invariant to extrinsic camera calibration inaccuracies and real-world deployment variations.

%% file: sections/3_4_defining_success.tex
\subsection{Defining Success.} 
\input{figures/5_env_and_pruner}

\label{sec:meth:sim:success}
For a given pruning cutpoint on a branch, we consider an end-effector pose to be successful if it satisfies both the orientation and reaching requirements. This means following a trajectory that avoids branches to move sufficiently near the cutpoint and changing the roll, pitch, and yaw of the end-effector to orient correctly for the cut. 
The reaching condition is satisfied if the end of the cutter is within 7~cm of the cutpoint along the branch~\cite{FlynnPruningDecision}, with the branch located within the cutter’s mouth. For orientation, two criteria must be met: the cutter must be \emph{pointing} toward the branch and must also be \emph{perpendicular} to it. Using Figure~\ref{fig:pointing_and_perp} (left) as a visual aid, to be correctly pointing, the cutter’s pointing vector ($\vec{v}_{\text{point}}$) should be perpendicular to the branch vector ($\vec{b}$), i.e., $\vec{v}_{\text{point}} \perp \vec{b}$. Due to practical constraints, we allow a tolerance of up to $30^\circ$ from perfect alignment~\cite{FlynnPruningDecision}, meaning any pointing vector within a blue cone of half-angle $30^\circ$ around the ideal direction --- as shown in Figure~\ref{fig:pointing_and_perp} (left) --- is acceptable. 
Similarly, to meet the perpendicularity requirement, the cutter's jaw orientation, defined by the vector $\vec{v}_{\text{perp}}$, should be parallel to the branch vector, i.e., $\vec{v}_{\text{perp}} \parallel \vec{b}$ (see Figure~\ref{fig:pointing_and_perp} (right)). Similar to the pointing requirement, we allow a  $30^\circ$ tolerance for the perpendicularity condition. Any vector within the illustrated triangle in Figure~\ref{fig:pointing_and_perp} (right) satisfies this requirement. The orientation condition is considered successful only if both the pointing and perpendicularity criteria are met.

 These criteria meet the requirements of an existing admittance controller that can guide the cutter once it makes contact with the branch~\cite{PruningAdmitanceController2022ICRA}. 

\input{figures/6_pointing_and_perp}

%% file: figures/5_env_and_pruner.tex
\begin{figure}
    \centering
    \includegraphics[width=\columnwidth]{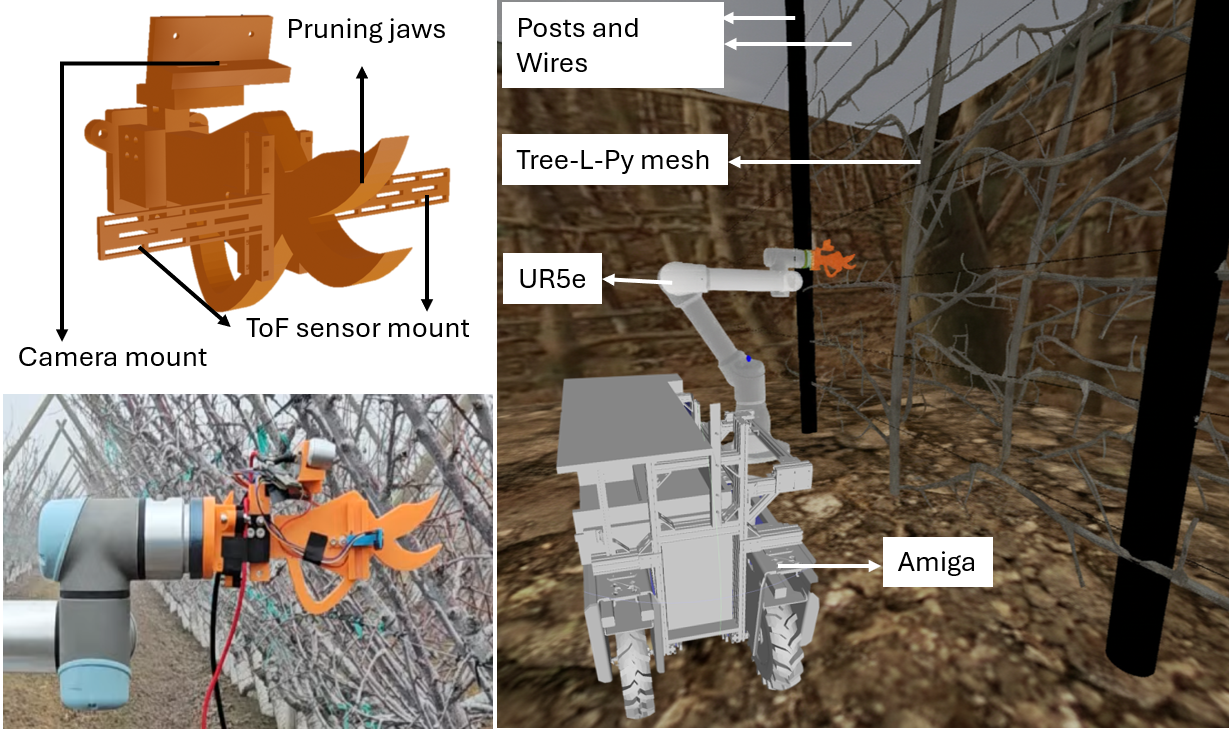}
    \caption{\textit{(Top left)} The model of the mock pruner. 
            \textit{(Bottom left)} The mock pruner attached to the UR5 arm. 
            \textit{(Right)} The simulation environment, including the robot and various orchard elements.}
    \label{fig:pruner_and_env}
\end{figure}

%% file: figures/6_pointing_and_perp.tex
\begin{figure}
    \centering
    \includegraphics[width=\columnwidth]{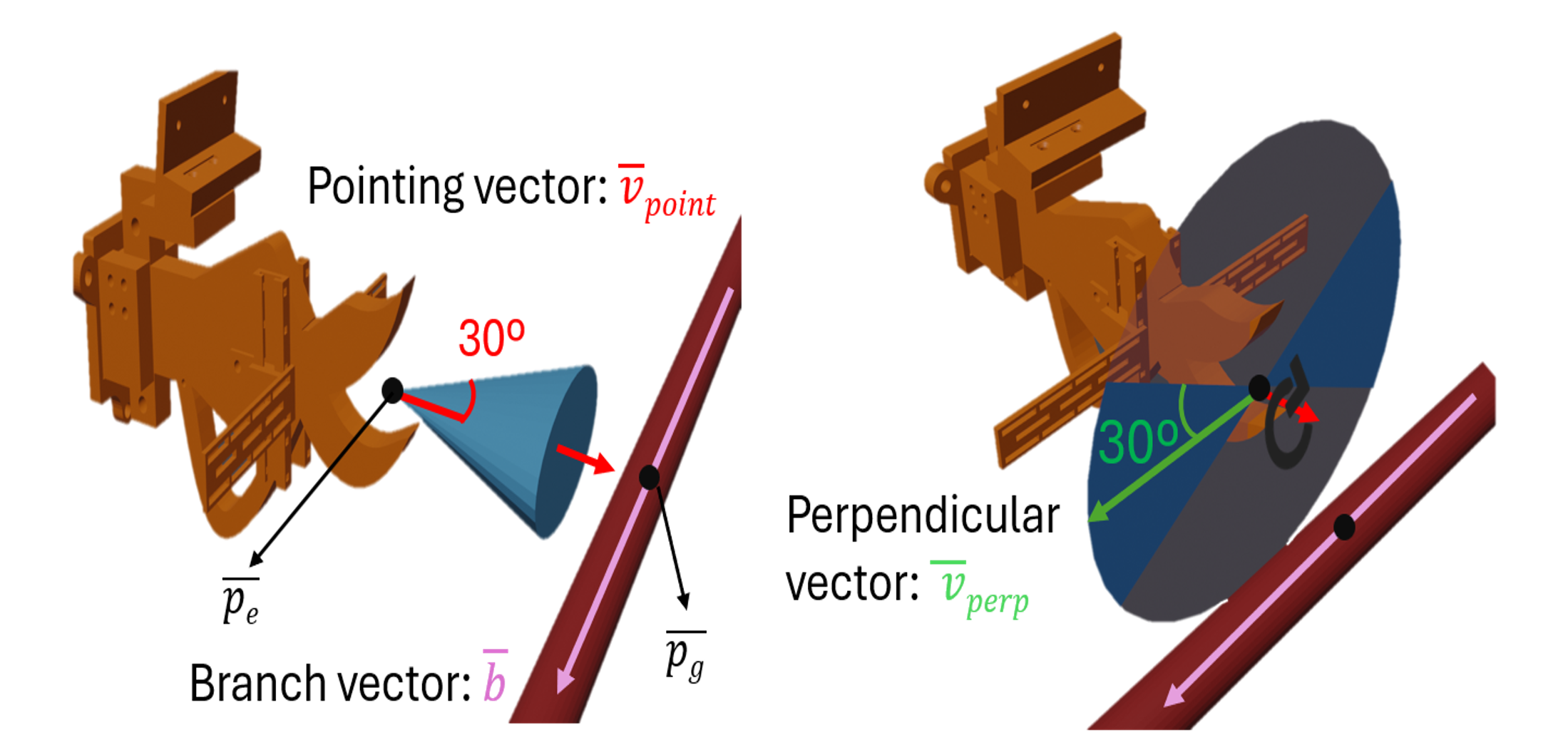}
    \caption{\textit{(Left)} To \emph{point}, $\vec{v}_{\text{point}} \perp \vec{b}$. A $30^\circ$ tolerance yields the set of feasible solutions shown as a blue cone. 
    \textit{(Right)} To be \emph{perpendicular}, $\vec{v}_{\text{perp}} \parallel \vec{b}$. A $30^\circ$ tolerance yields feasible solutions forming sectors of a circle.}
    \label{fig:pointing_and_perp}
\end{figure}

%% file: sections/4_0_setting_up_learning.tex
\section{LEARNING SETUP FOR VISUOMOTOR PRUNING}
\label{sec:meth:setting_up_learning}
We model vision-based control of a robotic pruner as a Partially-Observable Markov Decision Process (POMDP) defined by a state space $\mathcal{S}$, a continuous action space $\mathcal{A}$, environment transition dynamics $T: \mathcal{S}\times\mathcal{A}\rightarrow\mathcal{S}$, reward function $R: \mathcal{S}\rightarrow \mathbb{R}$, and observation function $\mathcal{O}: \mathcal{S} \rightarrow \Omega$ that maps states to partial observations $o \in \Omega$. We also have an offline dataset $\mathcal{D}_{\text{offline}}$ as a collection of trajectories $\tau = (o_0, a_0, r_0, o_1, \dots)$. Following standard practice in model-free reinforcement learning, we instantiate a parameterized policy $\pi_\theta: \mathcal{S} \rightarrow \mathcal{A}$ as a deep neural network that is optimized to maximize its expected discounted cumulative reward. The algorithm used here is \emph{Hybrid-PPO} (described in Section~\ref{sec:algorithm}) which incorporates an offline dataset along with online rollouts in simulation.

Achieving robust real-world deployment requires us to expose the policy to conditions it would face in the real-world. In this section, we first define our \textit{task space} and introduce a \textit{sampling procedure} to generate cutpoints that are sufficiently diverse. Next, we specify the \textit{state and action representations} used for learning and deployment. Finally, using the simulation environment with access to complete state information, we outline our method for \textit{automated collection of successful trajectories}.

%% file: sections/4_1_training_episodes.tex
\subsection{Generating Diverse Training Episodes}
\label{sec:meth:setting_up_learning:sampling}

An episode consists of a simulation scene, with a sampled cutpoint defined on one of the tertiary branches. On start of a new episode, the lighting, robot, and camera placement are randomized as outlined in section~\ref{sec:meth:sim}.\ref{sec:meth:sim:sim_setup}.

% When generating training episodes, to expose the agent to cutpoints those it might encounter in deployment, we uniformly sample from the workspace and the orientation of the corresponding to-be-pruned branch.
%to ensure the policy is exposed to sufficient conditions to cover those it might encounter in deployment. 
\vspace{4pt}\noindent\textbf{Task-space}
We define a \emph{likely}-reachable region for the UR5e robot which includes all points 75~cm to 105~cm from the robot base, not including those behind the vertical plane of the robot's base. This region is defined heuristically based on the robot's maximum reach with the cutters (125~cm), the maximum outward-facing branch length (40~cm), and the visibility of the tree for obstacle avoidance (closer than 30~cm yielded poor views). This region would allow the robot to reach pruning points on up to 4 trellis wires when placed in front of the tree in an orchard.\looseness=-1

\vspace{4pt}\noindent\textbf{Sampling procedure} To obtain a new cutpoint for each episode, we define a sampling procedure that uniformly covers the task space in both branch pose and spatial location, ensuring a diverse set of cutpoints.
First, we sample a cutpoint position uniformly within the \emph{task-space} along with a random 3D orientation \cite{graphic_gems_3}. Given a bank of 1,000 Tree-L-Py–generated trees, we search for tertiary branches whose orientation matches the sampled orientation within a tolerance of $\pm5^\circ$, and randomly select one such branch. The corresponding tree is then translated so that the selected branch aligns with the sampled position. This translation is constrained by the tree geometry to avoid unnatural configurations (e.g. tree floating mid air or the camera looking at empty spaces). Specifically, translation is limited to half the tree’s width in the left-right direction and half its height downwards, with resampling if these limits are exceeded. 
\looseness=-1

%% file: sections/4_2_state_action.tex
\subsection{Observation and Action Spaces}
\label{sec:meth:setting_up_learning:state_space}
While the simulator provides full information about tree geometry and pose, the policy observation space is limited to 1) a goal that specifies the cutpoint, 2) optical flow imagery, and 3) robot proprioception. A summary of the state space is shown in Figure~\ref{fig:state_space}.

\input{figures/7_state_space}

\vspace{4pt}\noindent\textbf{Cutpoint Specification.} To simulate real-world uncertainty during manipulation, we add uniform noise to the 3D cutpoint. Noise is sampled independently along each axis from a uniform distribution in the range $[-0.01, 0.01]$\,m, forming a small cube of possible perturbations centered at the cutpoint. This information is conveyed in two forms:  
(i) the noisy $(x, y, z)$ coordinates of the cutpoint relative to the robot’s end-effector, and  
(ii) a 1-channel \emph{cutpoint mask} ($424 \times 240$), where the noisy cutpoint is projected into camera space as a disk. The radius of this disk is inversely proportional to the distance of the cutpoint from the camera, keeping the apparent size constant.

\vspace{4pt}\noindent {\bf Optical Flow Image.}  Rather than providing the policy with rendered RGB or depth images, we instead use an off-the-shelf optical flow model RAFT~\cite{raft_of} to calculate a 2-channel $424\times240$ optical flow image. This image provides the change in 2D image coordinates per pixel between rendered RGB images from the current and previous time steps. Prior work has shown that this approach offers strong sim-to-real generalization~\cite{OpticalFlowAlex2022IROS}. Figure~\ref{fig:optical_flow_comparison} is an example of optical flow in both sim and real-world. In real-world orchards, we have verified that optical flow captures the geometry of small branches better than a depth camera as seen in Figure~\ref{fig:of_vs_depth}
\looseness=-1

\vspace{4pt}\noindent {\bf Proprioception.} We include the 6-DOF pose, the end-effector’s current velocity, and the robot’s joint angles in the state space. The end-effector’s pose is defined by its location
as $(x, y, z)$ coordinates and its orientation by 6D parameterization~\cite{6drotation}. The joint angles are represented using their sine and cosine values. This results in a 27-dimensional proprioception state.
\looseness=-1

\vspace{4pt}\noindent {\bf Actions.} The policy outputs linear and angular velocities in the end-effector frame. Joint velocities are then computed by using the damped least-squares controller and applied for 0.5 seconds --- corresponding to a control frequency of 2 Hz.

\input{figures/of_vs_depth}

%% file: figures/7_state_space.tex
\begin{figure}
    \centering
    \includegraphics[width=\columnwidth]{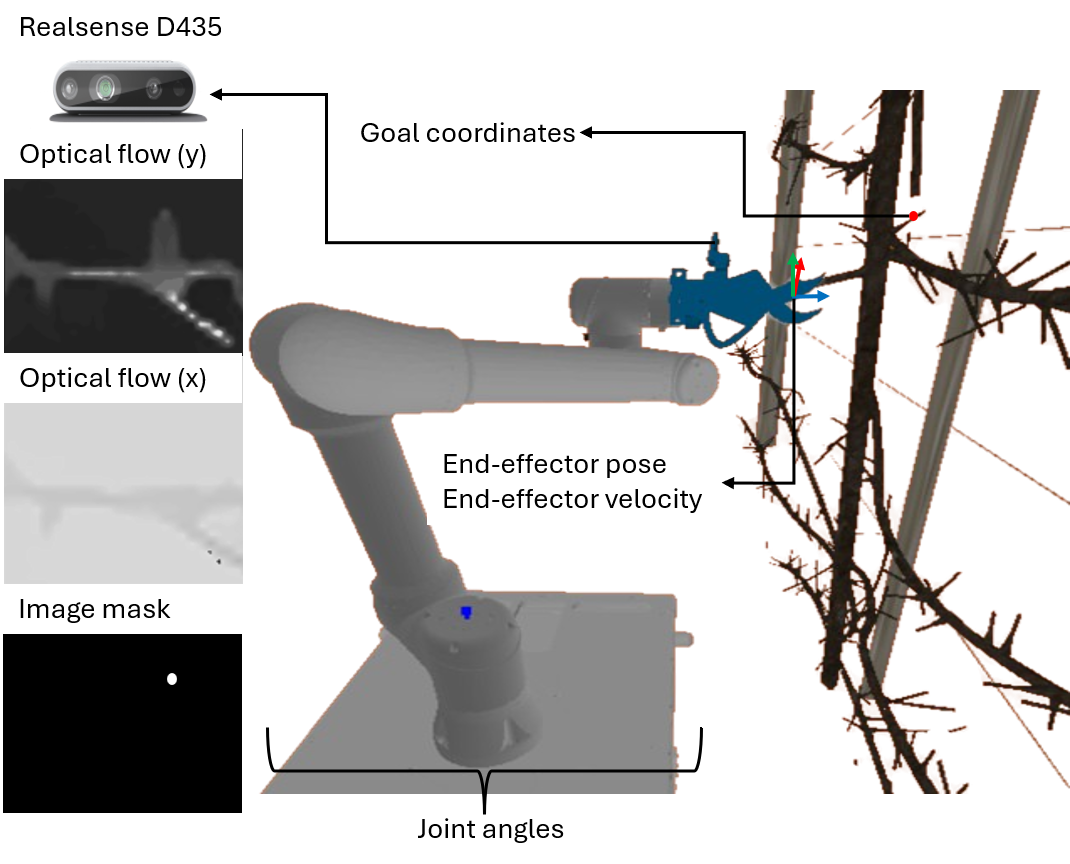}
    \caption{Our state space includes optical flow for perception and proprioceptive features, such as end-effector velocity and pose, joint angles, and the coordinates of the pruning goal.}
    \label{fig:state_space}
\end{figure}

%% file: figures/of_vs_depth.tex
\begin{figure}
    \centering
    \includegraphics[width=\columnwidth]{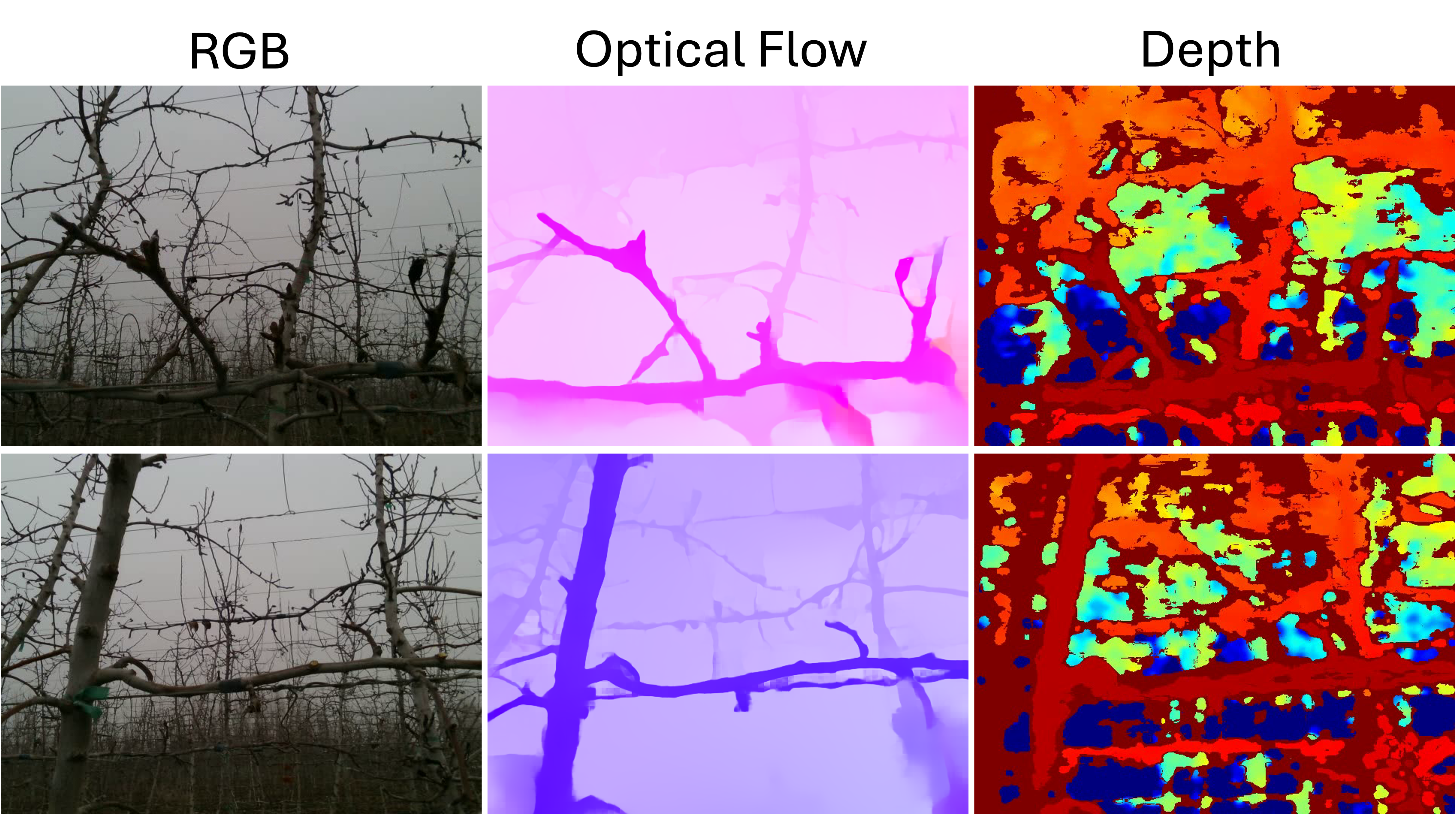}
    \caption{\textit{(Left)} RGB images of an orchard tree \textit{(Middle)} Corresponding optical flow image obtained using RAFT~\cite{raft_of} \textit{(Right)} Corresponding depth image using Intel RealSense D435 camera.}
    \label{fig:of_vs_depth}
\end{figure}

%% file: sections/4_3_policy.tex
\subsection{Policy Architecture}
\label{sec:meth:setting_up_learning:policy}
We consider a LSTM-based actor-critic architecture with the addition of a privileged critic that has access to training-time only features and an encoder for visual feature learning.

\vspace{4pt}\noindent \textbf{Visual encoder.} We stack the cutpoint mask and optical flow images into a single 3-channel $424 \times 240$ image $I_t$, and pass it through a convolutional encoder to obtain a compact latent representation $z_t \in \mathbb{R}^{64}$. The encoder consists of a 5-layer convolutional network with LayerNorm after each convolution and a ReLU non-linearity. The convolutional layers progressively downsample the input image to a feature map of size $8 \times 8 \times 14$, using strided convolutions. The final output is flattened and passed through a linear layer to yield a 64-dimensional embedding.

\vspace{4pt}\noindent \textbf{Actor.} The actor module takes the image embedding $z_t$ along with additional scalar observations: proprioceptive features, and the cutpoint representation. These are concatenated and passed through a 2-layer LSTM (hidden size 128), followed by a multi-layer perceptron (MLP) with hidden dimensions $[256, 128, 64]$ with LayerNorm and ReLU applied after each linear layer. An additional MLP without any LayerNorm and non-linearity outputs a 6-dimensional mean action vector corresponding to 6-DOF end-effector velocities. The final actions are modeled as $\tanh$-squashed Gaussians  \cite{SACTanh} with a shared learned standard deviation, and scaled to lie within $\pm 0.2\,\text{m/s}$.

\vspace{4pt}\noindent \textbf{Privileged Critic.} The critic shares the same image encoder as the actor and has an identical architecture, except: (i) it outputs a single scalar, (ii) it receives two additional scalar features corresponding to the pointing cosine similarity and perpendicular cosine similarity between the end-effector and the to-be-pruned branch. These inputs rely on privileged knowledge of the tree geometry and pose \cite{PrivilegedCriticPinto} but are not available to the policy and are only used during training. \looseness=-1

\input{figures/8_optical_flow_comparison}

%% file: figures/8_optical_flow_comparison.tex
\begin{figure}
    \centering
    \includegraphics[width=\columnwidth]{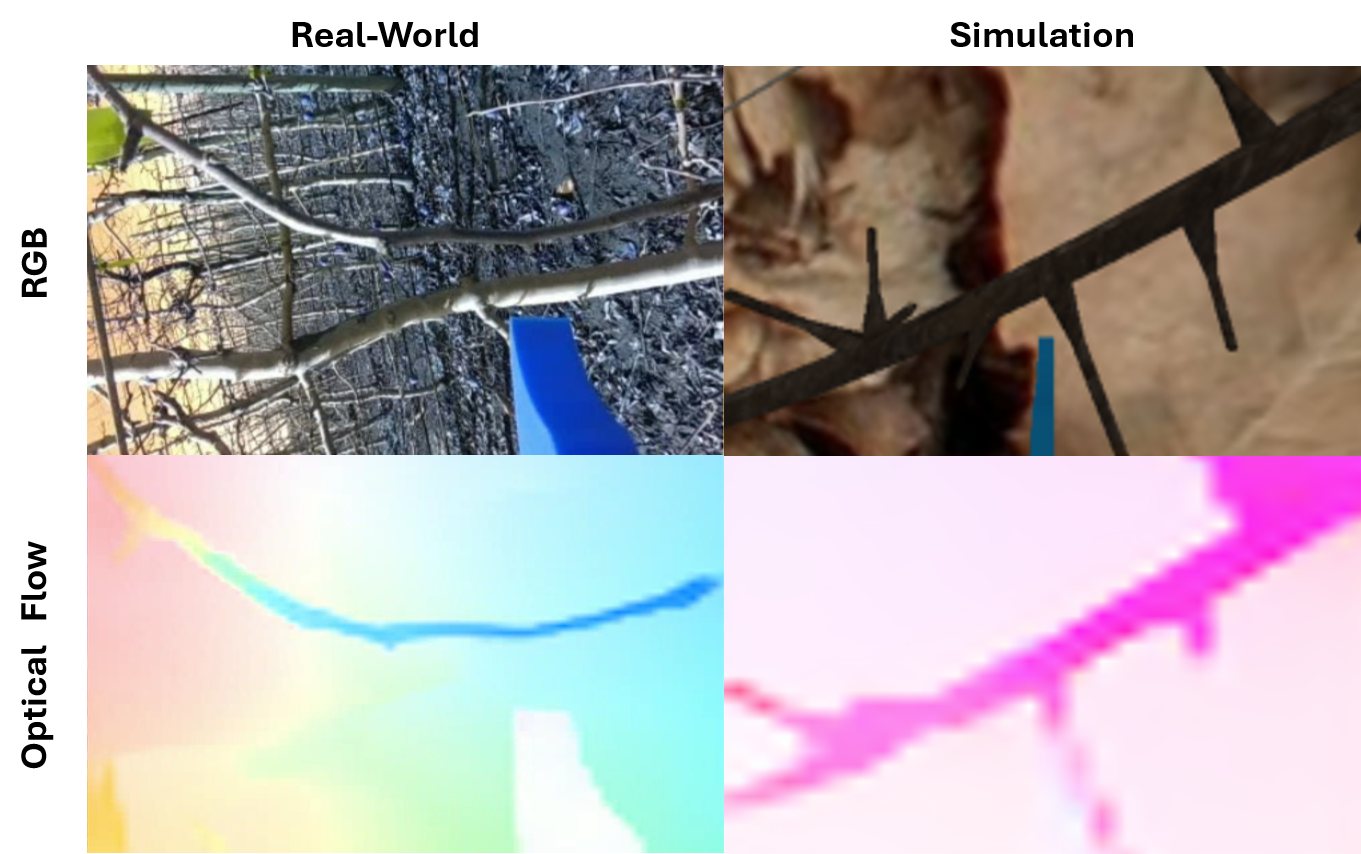}
    \caption{\textit{(Top)} RGB images from the real world (left) and simulation (right). 
        \textit{(Bottom)} Optical flow from the corresponding real world (left) and simulation (right) images. 
        The color differences arise from the different motions performed.}
    \label{fig:optical_flow_comparison}
\end{figure}

%% file: sections/4_4_rewards.tex
\subsection{Reward}
\label{sec:meth:setting_up_learning:reward}

Our reward function is designed to encourage \emph{reaching} towards the target, \emph{pointing} at the branch to be pruned, and aligning the end-effector to be \emph{perpendicular} to the branch. In addition, the reward includes a \emph{success bonus} when the branch enters the jaws of the cutter when aligned correctly, \emph{penalties for collisions}, and a \emph{slack term} to promote efficient motions.

Refer to Figure~\ref{fig:pointing_and_perp} for the notations used in defining the reward components. We represent the end-effector's rotation matrix at time \( t \), \( R(t) \), using the unit vectors \( \vec{x}, \vec{y}, \vec{z} \), which correspond respectively to the \emph{pointing} direction (\( \vec{v}_{\text{point}} \)), the \emph{perpendicular} direction (\( \vec{v}_{\text{perp}} \)), and the up-down axis. The branch growth direction is denoted by the vector \( \vec{b} \). The position of the cutter jaws' center at time \( t \) is given by \( p_e(t) \), and the desired cutpoint on the branch is represented by \( p_g \). All vectors are transformed to the robot base frame while calculating the rewards.

\looseness=-1

%Our reward function has a term to move the end effector closer to the goal point, two  orientation terms (one for pointing at the branch, the second for driving the cutter perpendicular to the branch), a hard and a soft collision penalty, a slack reward to encourage faster motion, and a one-shot positive reward when the branch crosses into the mouth of the cutter. The movement and orientation rewards are positive if the end-effector moved in the correct direction from the last time step, and negative if it moved the wrong direction.

\vspace{4pt}\noindent \textbf{Reaching Reward.} Let $p_e(t)$ be the point at the center of the cutter jaws at time $t$ and $p_g$ be the cutpoint. Then the \emph{reaching} reward at time $t$ is the change in distance between the cutpoint and end-effector at time $t$ and $t-1$:
\begin{equation}
R_{\mathtt{reach}}(t)  = \left\| p_{\mathrm{e}}(t{-}1) - p_{\mathrm{g}} \right\|_2 - \left\| p_{\mathrm{e}}(t) - p_{\mathrm{g}} \right\|_2
\end{equation}

\vspace{4pt}\noindent \textbf{Pointing Reward.} For orientation, we use cosine similarity ($C(v_1, v_2) \rightarrow [-1,1]$) to compare vectors. If the pointing vector $\vec{v}_{\text{point}}(t) = R(t)\vec{x}$ points at the branch, then the perpendicular projection of $p_g - p_e(t)$ on $\vec{b}$ denoted here as $\perpproj{(p_g - p_e(t))}{\vec{b}}$ should be in the same direction as $\vec{v}_{\text{point}}(t)$. As with the reaching reward, we provide the change in this cosine similarity as the pointing reward:
\begin{equation}
\begin{split}
    R_{\mathtt{point}}(t) = &\ C\left(\perpproj{({p}_{\mathrm{g}} - {p}_{\mathrm{e}}(t))}{\vec{b}}, \vec{v}_{\text{point}}(t)\right) \\
    &- C\left(\perpproj{({p}_{\mathrm{g}} - {p}_{\mathrm{e}}(t{-}1))}{\vec{b}}, \vec{v}_{\text{point}}(t{-}1)\right)
\end{split}
\end{equation}
\vspace{4pt}\noindent \textbf{Perpendicularity Reward.} The perpendicularity vector $\vec{v}_{\text{perp}}(t) = R(t) \vec{y}$ also needs to be aligned to the branch. As with prior rewards, we use the change in cosine similarity between $\vec{v}_{\text{perp}}(t)$ and $\vec{b}$ as our perpendicularity reward:
\begin{equation}
R_{\mathtt{perp}}(t)  = \left|C(\vec{v}_{\text{perp}}(t), \vec{b})\right| - \left|C(\vec{v}_{\text{perp}}(t-1), \vec{b})\right|
\end{equation}
\vspace{4pt}\noindent \textbf{Collision Penalties.} If the robot collides with the environment, there is a negative reward $R_{\mathtt{col}}$ of -0.01 for a small branch and -0.1 for other rigid structures like trunks, primary branches, posts, wires and itself. These semantic part labels are automatically derived from the mesh metadata generated during the synthetic tree growth process (Section~\ref{sec:meth:sim}).

\vspace{4pt}\noindent \textbf{Termination and Overall Reward.} We terminate an episode if 100 steps are reached or if the end-effector meets the success criteria for the given cutpoint. If successful, a terminal reward $R_{\mathtt{term}}$ of 3 is provided and 0 otherwise. The total reward at timestep $t$ is then the weighted sum of these individual rewards plus a constant slack $R_{\mathtt{slack}} = -0.1$ reward to encourage efficiency,
\begin{equation}
    \begin{split}
        R(t) = \alpha_{m} R_{\mathtt{reach}} + \alpha_{p1} R_{\mathtt{perp}} + \alpha_{p2} R_{\mathtt{point}}\\ + R_{\mathtt{term}} + R_{\mathtt{slack}} + R_{\mathtt{col}}    
    \end{split}
\end{equation}
\noindent where $\alpha_m = 5$, $\alpha_{p1} = 6$, $\alpha_{p2} = 3$ are empirically determined weighting coefficients.

%We weight the move and perpendicular terms ($\alpha_m = 5$, $\alpha_{p1} = 6$) higher than the pointing reward  ???

%% file: sections/4_5_synthetic_data.tex
\subsection{Generating Synthetic Data}
\label{sec:meth:setting_up_learning:synthetic_data}
Collecting a large dataset of successful pruning trajectories manually---on the order of thousands---would be prohibitively expensive due to the human effort required. To address this, we automate this generation process by leveraging our simulation environment with access to perfect state information. 

Specifically, we use RRT-Connect\cite{rrt_connect}, a sampling-based motion planning method that produces waypoints satisfying the success criterion for pruning. These waypoints are then post-processed into smoother trajectories and converted into transition tuples --- comprising observation, action, reward, and next observation --- to be compatible with the reinforcement learning setup.

\vspace{4pt}\noindent\textbf{Generating goal configuration.} 
Each pruning point can have multiple end-effector poses satisfying the success criterion based on the pointing direction and due to the tolerances present in reaching, pointing and perpendicularity requirements. While some of these poses are reachable via collision-free trajectories, others are infeasible due to collisions with surrounding orchard clutter. To identify feasible solutions, we generate a set of end-effector configurations that satisfy the pruning criterion and use RRT-Connect to compute collision-free trajectories to these poses.

To sample valid end-effector poses, we begin with the ideal end-effector pose --- one that is perfectly aligned to the to-be-pruned branch, closest in orientation to the initial end-effector frame and satisfying the success criterion. 

To introduce variation in the direction of approach, this ideal pose is rotated about the branch direction $(\hat{b})$ (see Figure~\ref{fig:pointing_and_perp}), with the rotation angle sampled from a Von Mises distribution centered at $0^\circ$. This biases the samples toward ideal alignment while maintaining diversity in the resulting set of goal poses. 

Next, the resulting pose is perturbed randomly around the pointing and perpendicular vectors. These perturbations are constrained within task-defined thresholds, ensuring all sampled goal poses fall within the acceptable range for successful pruning. Finally, each pose is converted into a robot configuration (joint states) using inverse kinematics to be passed to RRT-Connect.
 
 \vspace{4pt}\noindent \textbf{Generating trajectories.} For each pruning point, we run RRT-Connect by sampling goal configurations (as explained above) until a valid trajectory is found or until 100 attempts have been made, after which the point is discarded. Directly sampling goal poses on the branch surface often caused the planner to fail due to proximity to collisions. To mitigate this, we offset the goal pose by 5 cm along the branch normal—keeping the end-effector oriented toward the branch. Once a valid path is found, the final goal pose is appended as the last configuration. 
 
 RRT-Connect finds feasible paths composed of discrete robot configuration waypoints, these paths are not smooth and typically require post-processing\cite{efficient_path_rrt}. We apply shortcutting\cite{efficient_path_rrt}, a post-processing technique that samples random points along the path and attempts to directly connect them to form shorter, smoother, yet feasible paths. The resulting smoothed paths are then sub-sampled every 5 cm to create dense waypoint sequences. These dense waypoints are converted into trajectories by computing end-effector velocity commands at a control frequency of 2Hz. For each pair of consecutive waypoints, a velocity vector is computed and scaled to ensure the end-effector reaches the next point within the fixed control interval.

\vspace{4pt}\noindent \textbf{Creating the dataset.}
To generate transition tuples---comprising observation, action, reward, and next observation--- we replay the velocity action trajectories in simulation, recording transitions at each time step. Only trajectories that end in successful configurations are retained.

To construct the trajectory-level dataset, we uniformly sample pruning points across the workspace, as described in Section~\ref{sec:meth:setting_up_learning}.\ref{sec:meth:setting_up_learning:sampling}. In total, we generate 5,000 unique pruning points across 1,000 different trees and apply the synthetic data generation pipeline for each of UFO and V-Trellis. After filtering out unsuccessful attempts, the final dataset contains 4,282 successful pruning trajectories for V-Trellis and 4,424 for UFO.

%% file: sections/5_0_learning_algorithm.tex
\section{LEARNING A VISUOMOTOR PRUNING POLICY}
\label{sec:algorithm}
Proximal Policy Optimization (PPO)~\cite{Schulman2017ProximalPO} has been used successfully to train visuomotor pruning policies~\cite{JainRL}. Recent work~\cite{nair2020awac, qfilter, rajeswaran2018dapg, ramrakhya2023pirlnav} has shown that combining reinforcement learning with an offline dataset of successful trajectories can improve performance compared to using RL alone. Motivated by these findings, we augment the learning process using a synthetically generated dataset of successful trajectories.

The synthetic data generation pipeline (described in Section~\ref{sec:meth:setting_up_learning}.\ref{sec:meth:setting_up_learning:synthetic_data}) uses the full environment state available in simulation to generate trajectories, while the learned policy operates under partial observability. As a result, directly imitating such trajectories can lead to sub-optimal actions; for instance, the policy may begin rotating the end-effector before the relevant branch is visible.

Directly applying behavior cloning (BC) to this dataset reproduces such sub-optimal behaviors. Our goal, therefore, is to develop an algorithm that can stitch together offline and online trajectories, while efficiently moving away from poor actions present in the offline data. Methods such as PPO+BC~\cite{ppo_bc} can partially mitigate this issue by adding a behavioral cloning loss to online updates, but they do not explicitly reject sub-optimal actions and the competing losses can degrade performance as demonstrated by hybrid-RL algorithms often outperforming baselines that add a BC loss~\cite{qfilter, nair2020awac}.

In this section, we propose a Hybrid Proximal Policy Optimization (Hybrid-PPO) method 
that incorporates per-decision importance sampling to correct for off-policy data and integrate it into the PPO framework. Doing so allows us to leverage simulated rollouts and offline data concurrently. This advantage-weighted formulation enables the critic to reweight offline actions; by assigning lower advantage estimates to suboptimal actions, the policy avoids favoring these actions present in the offline dataset. In Section~\ref{sec:experiments:algo}, we validate the proposed 
algorithm by comparing it against an online-only PPO baseline, an offline-only BC baseline, and a simple PPO+BC baseline on the pruning task.

%% file: sections/5_1_problem_definition.tex
\subsection{Problem Definition}

We model learning a visuomotor control policy for pruning as a POMDP (Section ~\ref{sec:meth:setting_up_learning}). In addition we define an offline dataset $\mathcal{D}_{\text{offline}}$ as a collection of trajectories $\tau = (o_0, a_0, r_0, o_1, \dots)$ generated using full-state access ($s_t$) but paired with the corresponding partial observation $o_t$.  

The key challenge here lies in leveraging the offline dataset $\mathcal{D}_{\text{offline}}$ generated using full state to learn a policy $\pi_\theta(a_t \mid o_t)$ operating under partial observability, relying on inputs from the observation space $\mathcal{O}$. Since the full simulation state $(s_t)$ can be intractable, we denote an approximation to the behavior policy used to collect the data as $\mu(a_t \mid o_t)$ and our target policy $\pi_\theta(a_t \mid o_t)$. To learn appropriately, this requires accounting for the distributional shift between $\mu$ and $\pi_\theta$.

The learned policy $\pi_\theta$ is a squashed Gaussian policy parameterized by a neural network, where the mean is predicted conditioned on $o_t$, and the output is passed through a $\tanh$ function. The standard deviation is state-independent and a learned parameter.

%% file: sections/5_2_ppo.tex
\subsection{Proximal Policy Optimization}
\label{sec:algorithm:ppo}
We first review the standard Actor-Critic framework and the Proximal Policy Optimization (PPO) algorithm~\cite{Schulman2017ProximalPO}. 

\vspace{4pt}\noindent\textbf{Actor–critic methods.} They are a class of reinforcement learning algorithms that update the policy using low-variance gradient estimates obtained by the use of a learned value function. These methods maintain two separate function approximators: an \emph{actor}, which represents the policy $\pi_\theta(a \mid s)$ and selects actions, and a \emph{critic}, which estimates a value function $V_\psi(s)$ representing the expected reward obtained by the current policy from state $s$.

The actor is trained using policy gradient methods by maximizing the expected return:
\begin{equation}
    J(\theta) = \mathbb{E}_{\pi_\theta} \left[ \sum_{t=0}^{T} \gamma^t r_t \right],
\end{equation}
where $\gamma \in (0,1)$ is the discount factor. The policy gradient is given by:
\begin{equation}
    \nabla_\theta J(\theta) = 
    \mathbb{E}_t \left[ 
    \nabla_\theta \log \pi_\theta(a_t \mid s_t) \, A_t
    \right],
\end{equation}
where $A_t$ is the advantage function, which measures the relative benefit of executing an action $a_t$ as compared to the expected value ($V(s_t)$) at the current state.
The critic provides a low-variance estimate of this advantage, commonly using a learned value function:

\begin{equation}
    A_t = r_t + \gamma V_\psi(s_{t+1}) - V_\psi(s_t).
\end{equation}

\vspace{4pt}\noindent\textbf{PPO.} In vanilla actor-critic methods, the training pipeline collects data using the 
current policy, performs a single gradient update on the actor and critic, and immediately discards the batch to prevent off-policy instability. To improve sample efficiency, PPO enables multiple gradient steps on the same rollout batch 
by introducing an importance-sampled, clipped surrogate objective. This clipped probability ratio acts as a simple first-order trust region, keeping the updated 
policy close to the data collecting policy and preventing the high variance and training instability typical of repeated off-policy updates. The clipped objective is defined as:

\begin{align}
\label{eq:ppo_clip}
L^{\text{CLIP}}(\theta) =
\mathbb{E}_t \Big[
\min \big(
r_t(\theta) \, \hat{A}_t, \;
\text{clip}(r_t(\theta), 1 - \epsilon, 1 + \epsilon) \, \hat{A}_t
\big)
\Big],
\end{align}
where $r_t(\theta)$ is the importance sampling ratio between the updated policy  $(\pi_{\theta})$ and the data collecting policy $(\pi_{\theta_\text{old}})$:
\begin{equation}
\label{eq:clip_ratio}
r_t(\theta) = 
\frac{\pi_\theta(a_t \mid s_t)}{\pi_{\theta_{\text{old}}}(a_t \mid s_t)}.
\end{equation}

The advantage estimate $\hat{A}_t$ is computed using the Generalized Advantage Estimator (GAE)~\cite{gae}:
\begin{equation}
\label{eq:ppo_advantage}
\hat{A}_t = 
\sum_{l=0}^{\infty} 
(\gamma \lambda)^l 
\delta_{t+l},
\end{equation}
where $\delta_t$ is the one-step temporal-difference (TD) error:
\begin{equation}
\delta_t = r_t + \gamma V_\psi(s_{t+1}) - V_\psi(s_t),
\end{equation}
with $V_\psi(s_t)$ denoting the critic’s value estimate parameterized by $\psi$. 
The parameters $\gamma$ and $\lambda$ control the discount factor and bias–variance trade-off, respectively.

The critic (value function) is trained to predict the expected return from each state by minimizing the mean-squared error between its estimate and a bootstrapped return target
\begin{equation}
    \label{eq:ppo_val}
    L^{\text{value}}(\psi) =
    \mathbb{E}_t \Big[
    \big( V_\psi(s_t) - R_t \big)^2
    \Big],
\end{equation}
where the target return $R_t$ is defined as:
\begin{equation}
    \label{eq:ppo_returns}
    R_t = \hat{A}_t + V_\psi(s_{t}),
\end{equation}

%% file: sections/5_3_hybrid_ppo.tex
\subsection{Hybrid Proximal Policy Optimization (H-PPO)}
\label{sec:algorithm:hybrid_ppo}
When updating the policy using offline data, we seek to preserve the stability properties of PPO by bounding policy updates within a clipped importance sampling range $[1 - \epsilon, 1 + \epsilon]$. This constraint ensures stable learning even when incorporating transitions generated under a distinct behavior policy.

\vspace{4pt}
\noindent \textbf{Modeling the behavior policy.}
To estimate the behavior policy $\mu(a \mid o)$ that generated the offline dataset $\mathcal{D}_{\text{offline}}$, we train a density model via Behavior Cloning (BC)~\cite{bc}. 
This network shares the architecture of $\pi_\theta$ and maximizes the log-likelihood of offline actions:
\begin{equation}
    \mathcal{L}_{\text{BC}} 
    = \mathbb{E}_{(o_t, a_t) \sim \mathcal{D}_{\text{offline}}}
    \big[ \log \mu(a_t \mid o_t) \big].
    \label{eq:bc}
\end{equation}
The shared action variance of $\mu$ is treated as a hyperparameter. This behavioral model provides the baseline action probabilities required to compute importance sampling ratios.

\vspace{4pt}
\noindent \textbf{Computing off-policy advantage.} 
To address the distribution shift between the offline behavior policy $\mu$ and the target policy $\pi_\theta$, we employ a V-trace-inspired formulation~\cite{vtrace}. Per-decision importance sampling (IS) is applied to correct temporal-difference (TD) errors, while future IS ratios are truncated to bound variance.

We define the standard one-step TD error as:
\begin{equation}
    \delta_t^{\text{offline}} = r_t + \gamma V(o_{t+1}) - V(o_t),
\end{equation}
and the IS-adjusted TD error as $\delta_t^{\text{offline\_IS}} = \bar{\rho}_t \delta_t^{\text{offline}}$, where the unclipped IS ratio relative to the behavior policy is given by:
\begin{equation}
    \rho_t = \frac{\pi_{\theta_{\text{old}}}(a_t \mid o_t)}{\mu(a_t \mid o_t)}, \quad \text{and} \quad \bar{\rho}_t = \min(\bar{\rho}, \rho_t).
    \label{eq:is_ratio}
\end{equation}

These corrected errors are integrated into a modified Generalized Advantage Estimator (GAE). Because the advantage function $A^\pi(o_t, a_t)$ explicitly conditions on executing action $a_t$, evaluating the immediate transition requires no importance weight; the immediate reward and next-state distribution depend solely on environment dynamics given $a_t$. Importance sampling corrections are required only to align \textit{subsequent} trajectory steps with $\pi_\theta$.

The off-policy modified GAE is formulated as:
\begin{equation}
    \hat{A}^{\text{offline}}_t = \delta_t^{\text{offline}} + \sum_{l=1}^{T-t-1} (\gamma \lambda)^l \left( \prod_{k=0}^{l-1} c_{t+k} \right) \delta_{t+l}^{\text{offline\_IS}},
    \label{eq:hppo_advantage}
\end{equation}
where $c_t = \min(\bar{c}, \rho_t)$ represents the truncated trace coefficient, and $\bar{\rho}, \bar{c}$ are truncation thresholds used to control estimator variance.

\vspace{4pt}
\noindent \textbf{Actor update with IS correction.}
We modify the PPO actor loss to incorporate both the standard PPO clipping and the IS correction for offline data. For transitions sampled from $\mathcal{D}_{\text{offline}}$, the loss becomes:
\begin{align}
    \label{eq:hppo_clip}
    L^{\text{CLIP-offline}}(\theta) = 
    \mathbb{E}_{(o_t, a_t) \sim \mathcal{D}_{\text{offline}}} \Big[ \min\Big( &
    \rho_t \, r_t(\theta) \, \hat{A}_t^{\text{offline}}, \notag  \\
    \rho_t \, \text{clip}(r_t(\theta), 1 - \epsilon, 1 + \epsilon) \, \hat{A}_t^{\text{offline}} \Big)
    \Big].
\end{align}
Here, $\rho_t$ is the importance sampling ratio defined in Equation~\ref{eq:is_ratio}, whereas $r_t(\theta)$ is the clip ratio $
\frac{\pi_\theta(a_t \mid o_t)}{\pi_{\theta_{\text{old}}}(a_t \mid o_t)}$.
During training, each minibatch contains an equal split of online simulated rollouts ($\mathcal{D}_{\text{online}}$) and offline transitions ($\mathcal{D}_{\text{offline}}$). For online data, parameters are updated using standard PPO actor and critic losses (Eq.~\ref{eq:ppo_clip} and Eq.~\ref{eq:ppo_val}). For offline data, the actor is updated via the off-policy clipped objective (Eq.~\ref{eq:hppo_clip}) using the IS-corrected advantage estimates (Eq.~\ref{eq:hppo_advantage}). To prevent numerical instability from out-of-distribution actions, we clamp offline action log-densities to a minimum floor ($-20$). Additionally, a linear learning rate decay is applied across training. The overall training procedure is outlined in Algorithm~\ref{alg:hybrid-ppo}.

\begin{algorithm}[t]
\caption{Hybrid Proximal Policy Optimization (H-PPO)}
\label{alg:hybrid-ppo}
\begin{algorithmic}[1]
\State \textbf{Input:} Offline dataset $\mathcal{D}_{\text{dataset}}$
\State Initialize policy $\pi_\theta$, value function $V_\psi$, behavior policy $\mu$, and shared variance $\sigma$
\State Pretrain $\mu$ via behavior cloning and $V_\psi$ on discounted Monte Carlo return
\State Initialize $\pi_\theta \leftarrow \mu$

\For{each training iteration}
    \State Sample offline buffer $\mathcal{D}_{\text{offline}} \subset \mathcal{D}_{\text{dataset}}$
    \State Collect online rollouts into $\mathcal{D}_{\text{online}}$
    \While{data remaining in $\mathcal{D}_{\text{offline}}$ and $\mathcal{D}_{\text{online}}$}
        \State Sample offline batch $\mathcal{B}_{\text{offline}} \sim \mathcal{D}_{\text{offline}}$
        \State Sample online batch $\mathcal{B}_{\text{online}} \sim \mathcal{D}_{\text{online}}$
        
        \For{each sample $(o_t, a_t, r_t, o_{t+1}) \in \mathcal{B}_{\text{online}}$}
            \State Compute GAE $\hat{A}_t$ (Eq.~\ref{eq:ppo_advantage})
            \State Compute returns $R_t$ (Eq.~\ref{eq:ppo_returns})
        \EndFor
        
        \For{each sample $(o_t, a_t, r_t, o_{t+1}) \in \mathcal{B}_{\text{offline}}$}
            \State Compute V-trace GAE $\hat{A}_t^{\text{offline}}$ (Eq.~\ref{eq:hppo_advantage})
        \EndFor
        
        \State Normalize advantages:
        \Statex \qquad $\hat{A}_t, \, \hat{A}_t^{\text{offline}} \leftarrow \text{normalize}(\hat{A}_t, \, \hat{A}_t^{\text{offline}})$
        
        \For{$n = 1$ to $N_{\text{epochs}}$}
            \State Compute loss $\mathcal{L}^{\text{actor-online}}$ (Eq.~\ref{eq:ppo_clip})
            \State Compute loss $\mathcal{L}^{\text{actor-offline}}$ (Eq.~\ref{eq:hppo_clip})
            \State Compute critic loss $\mathcal{L}^{\text{critic}}$ (Eq.~\ref{eq:ppo_val})
            
            \State Combine actor losses:
            \Statex \qquad $\mathcal{L}^{\text{actor}} = \mathcal{L}^{\text{actor-online}} + \mathcal{L}^{\text{actor-offline}}$
            
            \State Total objective:
            \Statex \qquad $\mathcal{L}^{\text{total}} = \mathcal{L}^{\text{critic}} - \mathcal{L}^{\text{actor}}$
            
            \State Update $\pi_\theta, V_\psi$ by minimizing $\mathcal{L}^{\text{total}}$
            \State Update shared variance $\sigma$ maximizing $\mathcal{L}^{\text{actor-online}}$
        \EndFor
        \State Update learning rate
    \EndWhile
\EndFor
\end{algorithmic}
\end{algorithm}

%% file: sections/6_training.tex
\section{TRAINING}
\label{sec:training}
All training was conducted on an NVIDIA Tesla V100 GPU workstation. The Hybrid-PPO implementation was built upon the Stable Baselines3 framework~\cite{sb3}. We ran 20 parallel environments to perform online rollouts concurrently. The visuomotor pruning policy was trained for $2\text{M}$ online timesteps, using $4{,}282$ and $4{,}424$ synthetically generated offline trajectories corresponding to the V-Trellis and UFO orchard layouts, respectively. Training required approximately 4 days of wall-clock time.

The learning rate was set to $1\times10^{-5}$ for both the actor and critic networks, and $1\times10^{-4}$ for the shared variance. Importance sampling truncation thresholds were set to $\bar{\rho} = 1$ and $\bar{c} = 0.95$ to stabilize corrections from offline data. The policy variance was initialized at $0.2$, whereas the behavioral policy variance was set to $0.1$.

Policy checkpoints were saved every 75,000 steps. Checkpoints produced during the final 25\% of training were evaluated on a held-out test set of 150 pruning points sampled uniformly from previously unseen trees. The policy achieving the highest success rate on this test set was selected as the final model. The final selected policies, \textit{during training}, achieved a mean reward of $5.93$ with a success rate of $59\%$ on V-Trellis trees, and a mean reward of $6.1$ with a $57\%$ success rate on UFO trees in simulation.

%% file: sections/7_deployment.tex
\section{DEPLOYMENT}
\label{sec:deployment}
\noindent\textbf{Policy deployment}
The trained policy was deployed on an NVIDIA RTX~2080~TI system for real-time operation at a control frequency of 2Hz. The robot, comprising the UR5e arm, Amiga and the custom pruner was powered by an onboard battery system. ROS~2 \textit{Humble}~\cite{ROS} and \textit{MoveIt 2}\cite{moveit2} were used to control the robot, with the Time-of-Flight (ToF) sensors interfacing via \textit{MicroROS}~\cite{microROS}. 

For experiments, high-level decision-making and task sequencing were implemented using a behavior tree architecture built with PyTreesROS~\cite{py_trees_ros}. User interaction and manual overrides were provided through a PlayStation (PS) controller integrated with the behavior tree for ease of operation during field tests.

\vspace{1ex}\noindent\textbf{Field Setup:} 
For the physical evaluations, target pruning points were constrained to the bottom three trellis wires in the commercial UFO and V-Trellis orchards, whereas the laboratory setup utilized only the bottom two wires.

To define the robot's pose relative to the canopy, we establish a Global Canopy Reference Frame $\mathcal{F}_G = \{X_G, Y_G, Z_G\}$, where $Z_G$ points vertically along the support posts, $Y_G$ aligns with the orchard row, and $X_G$ extends normal to the fruiting wall.

For field trials, the mobile base was positioned to move along the row $Y_G$. The base was placed approximately $80\text{~cm}$ from the tree plane along $X_G$, with the end-effector pointing orthogonal to the row toward the canopy $X_G$ (Fig.~\ref{fig:robot}). This setup mirrors the geometric assumptions of our simulated task space. Although global positioning was executed manually for these trials, full automation of this navigation step is highly feasible by using orchard specific SLAM~\cite{treeslam} and image segmentation methods~\cite{treeseg} to align roughly orthogonal to the tree rows. The outdoor field trials were conducted over a two-day period under varying natural illumination conditions, ranging from overcast to sunny. These different conditions are shown in Figure~\ref{fig:field_weather}.

%% file: sections/8_0_experiment_setup.tex
\section{EXPERIMENTAL SETUP}
\label{sec:experiments}
We design our experimental evaluation to answer four core questions:
\begin{enumerate}[label=(\alph*)]
    \item \textbf{Algorithmic Validation:} How does our proposed Hybrid-PPO algorithm compare against Behavior Cloning (BC)~\cite{bc}, standard PPO~\cite{Schulman2017ProximalPO}, and PPO+BC~\cite{ppo_bc} baselines on simulated V-Trellis structures?
    \item \textbf{Policy Evaluation across Architectures in Simulation:} In exhaustive and diverse simulation trials, what is the overall performance of the learned policy across both V-Trellis and Upright Fruiting Offshoots (UFO) canopy architectures?
    \item \textbf{Real-World Sim-to-Real Transfer:} Can the trained policies transfer zero-shot to physical commercial V-Trellis and experimental UFO orchards under operational field conditions?
    \item \textbf{RRT-Connect Baseline Comparison:} How does our RL policy perform against a classical RRT-Connect motion planning baseline on physical hardware in a V-Trellis tree in laboratory setting?
\end{enumerate}

%% file: sections/8_1_algorithm_validation.tex
\subsection{Algorithm Validation}
\label{sec:experiments:algo}

To evaluate the proposed Hybrid-PPO algorithm, we compare it against Behavior Cloning (BC)~\cite{bc}, standard PPO~\cite{Schulman2017ProximalPO}, and PPO+BC~\cite{ppo_bc} baselines for the pruning task on the V-Trellis tree architecture in simulation. Each baseline is carefully selected to isolate a specific learning paradigm:

\begin{description}[font=\normalfont\itshape]
    \item[Standard PPO:] Relies solely on online rollouts performed in simulation (as described in Section~\ref{sec:algorithm}.~\ref{sec:algorithm:ppo}). This reflects the performance of a purely online method with no access to the offline dataset.
    
    \item[Behavior Cloning (BC):] Utilizes only the offline data in a supervised setting by maximizing Eq.~\ref{eq:bc}. This represents a purely offline method with no simulation interaction.
    
    \item[PPO+BC:] Incorporates offline data through a behavior-cloning loss term alongside online learning using PPO. This illustrates how our proposed algorithm performs relative to the simplest, most common hybrid approach. The resulting objective maximized is:
    \begin{equation}
        L^{\text{PPO+BC}}(\theta) = L^{\text{CLIP}}(\theta) + \lambda_{\text{BC}} L^{\text{BC}}(\theta)
        \label{eq:ppo-bc}
    \end{equation}
    where $L^{\text{CLIP}}(\theta)$ and $L^{\text{BC}}(\theta)$ are defined in Eq.~\ref{eq:ppo_clip} and Eq.~\ref{eq:bc}, respectively, with the weighting coefficient $\lambda_{\text{BC}}$ set to $0.005$.
    
    \item[Hybrid-PPO (H-PPO):] Our proposed algorithm, which integrates both online rollouts and offline data within a unified framework. A comprehensive description of this approach is detailed in Section~\ref{sec:algorithm}.\ref{sec:algorithm:hybrid_ppo}.
\end{description}
\vspace{1ex}\noindent\textbf{Training and Evaluation.}
For each algorithm, three policies were trained using three random seeds. Policy checkpoints were saved every 75,000 simulation steps. For each run, the final policy was selected from the last 25\% of checkpoints based on performance on a test set of 150 pruning points sampled uniformly from previously unseen trees, resulting in three final policies per algorithm. Each final policy was then evaluated on a set of 2,000 validation points from unseen V-Trellis trees. The BC baseline achieved a success rate below 5\% and is therefore excluded from further analysis.

\vspace{1ex}\noindent\textbf{Hierarchical Bootstrapping.}
To quantify statistical uncertainty in pruning success rates obtained by each algorithm, we employ hierarchical bootstrapping with two nested sources of randomness: (i) variation across training seeds and (ii) variation across evaluation samples. Using the three trained policies for each algorithm and their evaluations on 2,000 unseen points, we generate 10,000 bootstrap replicates of the success rate to obtain 95\% confidence intervals (CIs). We also perform pairwise comparisons by treating evaluation points as paired across methods and computing the bootstrap distribution of the mean differences between each baseline and Hybrid-PPO. Differences are considered \textit{statistically significant} if their CIs exclude zero.

%% file: sections/8_2_policy_eval.tex
\subsection{Policy Evaluation across Architectures in Simulation}
\label{sec:experiments:policy}

To evaluate the performance of our trained policy within our workspace, we conduct exhaustive \textit{simulation trials} with pruning points uniformly sampled across both location and orientation for both V-Trellis and UFO architectures. We also run an oracle RRT-Connect planner equipped with perfect environment information and a large computational budget to establish the upper bound on success rate. These comprehensive simulation experiments allow us to perform exhaustive evaluation across all possible branch configurations that would be logistically prohibitive and would pose hardware risks to perform in physical field trials.

For these evaluations, we generated 3,000 evaluation episodes using 100 previously unseen trees across both UFO and V-Trellis architectures. To achieve uniform coverage, cutpoints were generated by sampling 1,000 orientations, identifying the corresponding branches, and translating each to three distinct locations within the task space per orientation (described in Section~\ref{sec:meth:setting_up_learning}.\ref{sec:meth:setting_up_learning:sampling}). For each cutpoint, both the trained \textit{reinforcement learning policy} operating under partial observations and the \textit{oracle RRT-Connect planner} with access to the true mesh were executed. To measure errors, the closest ground-truth end-effector poses were recorded at trial completion. To compute these poses, starting from the final tool pose, the end-effector was positioned in front of the cutpoint, oriented toward the branch, and adjusted in yaw to align orthogonally to the branch—computed via the ground-truth branch pose in the simulator.

\vspace{1ex}\noindent\textbf{Oracle RRT-Connect Planner.} 
The RRT-Connect planner serves as an oracle in this context because it utilizes the perfectly accurate 3D tree mesh and is provided with a large computational budget. Specifically, the planner was run for 100 candidate goal configurations per cutpoint with up to 10,000 sampling steps for each goal. These goal configurations were sampled within the success region (see Section~\ref{sec:meth:setting_up_learning}.\ref{sec:meth:setting_up_learning:synthetic_data}) and offset to a pose 5~cm away, such that simply moving the robot forward would place the branch within the pruner. We consider the trial a success if a collision-free path exists to any of these sampled goals. We utilize the RRT-Connect implementation provided by PyBullet Planning~\cite{pybullet_planning}.

Obtaining such precise, noise-free 3D meshes in the real world is extremely difficult, making this setup unrealistic for practical field applications; however, this rigorous evaluation establishes the absolute \emph{best-case performance} limit for a classical planner given perfect perception.

\vspace{1ex}\noindent\textbf{RL Policy.} 
We executed the learned RL policy by performing the most likely action predicted by the network. For each cutpoint, similar to how it was trained, the policy was run for 100 simulation steps at 2~Hz or until the termination condition was reached.

%% file: sections/8_3_real_world_eval.tex
\subsection{Real-World Sim-to-Real Transfer}
\label{sec:experiments:real_world}

To validate the sim-to-real transfer of our learned policies, we evaluate our system across $38$ real-world trials split between two settings: (i) $10$ controlled laboratory trials on a constructed V-Trellis setup, and (ii) $28$ outdoor field trials across commercial V-Trellis apple and experimental UFO cherry orchards under operational field conditions.

\input{figures/field_weather}

The orchards where the experiments were performed are located in Prosser, WA, USA. The outdoor field trials were conducted under diverse environmental and atmospheric conditions, including overcast, foggy skies, mild rain, partly cloudy conditions, and direct, bright sunlight and can be seen in Figure~\ref{fig:field_weather}.  

Due to the distinct geometries of the V-Trellis and UFO architectures, separate policies were trained and deployed for each. Whether joint training could leverage shared structural representations across canopy types remains to be investigated.

For these trials, the cutpoints were selected within the reachable workspace and biased toward safer configurations that matched our training assumptions. Specifically, points on branches longer than $12.5$~cm were selected, employing a simplified decision-making criterion for branch selection~\cite{FlynnPruningDecision}. Target branches were chosen to ensure diversity in both spatial location and orientation while remaining strictly within the manipulator's workspace envelope. Because the orchard trees were grown at an incline (except the vertical laboratory setup), the reachable workspace was shifted, rendering branches inside the canopy physically inaccessible. Consequently, target selection was biased toward the more exposed outer canopy; as a result, the reported success rates likely represent an upper bound on generalized field performance.  Nevertheless, the primary goal of the real-world trials is not to benchmark performance, but to demonstrate the successful sim-to-real transfer of our learned policy.

\vspace{1ex}\noindent\textbf{RL Policy Experiments.} 
In the real-world experiments, a trial was considered complete when the execution time exceeded the simulated training horizon, the success criteria were met, or an unacceptable collision was anticipated. Specifically, a trial was terminated if: (a) a maximum operational duration of 80 seconds had elapsed; (b) the target branch was successfully aligned within the cutting mechanism (i.e., both time-of-flight sensors registered $\leq 7~\text{cm}$) and the end-effector was positioned within 7~cm of the designated cutpoint along the branch; or (c) the operator manually intervened to prevent an imminent hardware collision. These termination conditions were explicitly designed to mirror the simulated training scenarios and were evaluated manually by the operator during physical deployment.

% \vspace{1ex}\noindent\textbf{RRT-Connect Baseline Experiments.} 

%% file: figures/field_weather.tex
\begin{figure}
    \centering
    \includegraphics[width=\columnwidth]{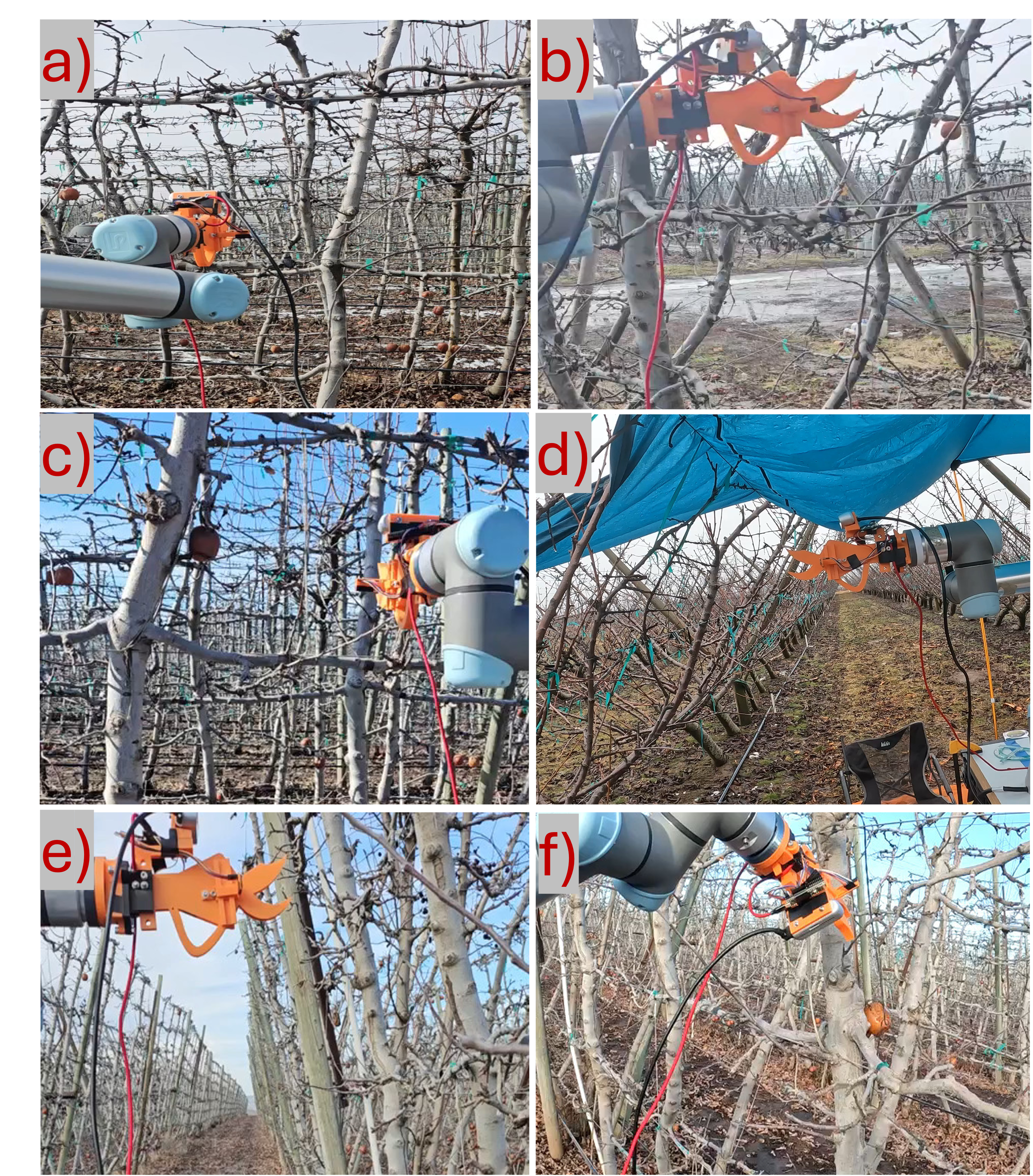}
   \caption{Demonstration of the autonomous pruning robot operating under diverse weather and lighting conditions: (a)~overcast, (b)~foggy, (c)~bright sunlight, (d)~rain with an artificial tarp canopy, (e)~shaded, and (f)~clear outdoor conditions.}
    \label{fig:field_weather}
\end{figure}

%% file: sections/8_4_real_world_baseline.tex
\subsection{RRT-Connect Baseline Comparison}
\label{sec:experiments:real_world_baseline}

To establish a physical hardware baseline, we implement a classical RRT-Connect motion planner using MoveIt~\cite{moveit2} and OMPL~\cite{ompl}. Both the learned policy and RRT-Connect were evaluated on a constructed V-Trellis tree in a controlled indoor laboratory environment across $15$ target cutpoints. Conducting baseline evaluations in the laboratory rather than the field was necessitated by limited outdoor testing time. Targets were sampled across diverse locations and orientations without biasing toward safer configurations—comprising $8$ and $7$ points across the lower and upper trellis levels, respectively.

By conducting these baseline experiments indoors, we provided the classical planner with a more reliable perception scenario. In outdoor orchards, the depth estimation of standard cameras (e.g., Intel RealSense D435) degrades significantly due to sunlight interference~\cite{lidaroutdoorbad}. Therefore, this laboratory evaluation represents an upper-bound performance for the RRT-Connect baseline on physical hardware. 

To execute the RRT-Connect planner, we first construct a 3D mesh representation of the tree structure to serve as collision geometries for motion planning. This 3D-reconstruction pipeline to obtain the mesh and the setting of the planner are described below.

\vspace{1ex}\noindent\emph{3D-Reconstruction Pipeline:} 
To provide the necessary collision map for RRT-Connect, we built a mesh of the tree using a custom RGB-D point cloud fusion pipeline. Relying solely on depth measurements proved insufficient due to the uniform, repetitive geometry of the branches; therefore, utilizing full RGB-D data was essential to leverage color for spatial disambiguation. 

\vspace{1ex}\noindent\textit{Capturing Views:} 
Capturing the tree geometry presented a unique spatial challenge: scanning from too far away missed critical small branches that require pruning, whereas close-up views lacked sufficient geometric features for point cloud stitching. Based on the visual fidelity of the resulting point clouds, we determined that a distance of $80\,\text{cm}$ between the camera and the tree provided optimal RGB-D views for 3D reconstruction. To capture the entire tree canopy from this distance, we executed a ladder-like scanning trajectory, acquiring $12$ distinct poses (Fig.~\ref{fig:3d_reconstruction}) across the bottom two trellis wires. During processing, raw depth measurements were strictly truncated at $1.0\,\text{m}$ to reject background environmental clutter and isolate the target canopy.
Initial spatial alignment for each capture was established using the robot arm's highly accurate forward kinematics. 

\vspace{1ex}\noindent\textit{Stitching the Point Cloud:} 
To fuse the individual point clouds into a complete tree mesh, we utilized a sequential multi-scale colored ICP algorithm~\cite{ZhouOpen3d} to iteratively align each new frame to a continuously growing global map. Registration employed a coarse-to-fine hierarchy across three voxel downsampling radii ($4\,\text{cm}$, $2\,\text{cm}$, and $1\,\text{cm}$). The larger radii facilitated coarse alignment of major structural elements, such as the main trunk and trellis wires, while subsequent finer iterations aligned smaller branches, spurs, and structural wires. Following fusion, the global point cloud was downsampled to a final $5\,\text{mm}$ voxel resolution and filtered via statistical outlier removal to eliminate residual sensor noise. Finally, this point cloud (Fig.~\ref{fig:3d_reconstruction}) was converted into a 3D mesh using $\alpha$-shape reconstruction~\cite{edelsbrunner1994three} in Open3D and imported into MoveIt as a planning scene collision object.

\vspace{1ex}\noindent\emph{Execution:} 
Prior to trial execution, the robot base was positioned approximately $80\,\text{cm}$ from the tree to ensure the target canopy remained well within the manipulator's reachable workspace. From this home configuration, an initial depth scan was captured at an $80\,\text{cm}$ distance to rigidly align the pre-reconstructed global mesh with the robot's updated base frame. For each selected cutpoint, rather than computing goal poses dynamically, we manually supplied collision-free target end-effector poses offset by $5\,\text{cm}$ from the branch and oriented orthogonally. This $5\,\text{cm}$ offset was necessary because the planner consistently failed to find valid paths when target poses resided directly on the branch surface, interpreting the target branch itself as an environment collision in dense canopy clutter. The RRT-Connect planner was executed only for cutpoints successfully represented in the reconstructed mesh, with the motion planning search timeout capped at $60\,\text{s}$.

%% file: sections/9_0_experiments.tex
\section{RESULTS}
\label{sec:results}
This section presents the findings addressing each of our four research questions:
\begin{itemize}
    \item \textbf{Algorithmic Validation:} Hybrid-PPO achieves higher mean success rates compared to BC, standard PPO, and PPO+BC baselines.
    \item \textbf{Policy Evaluation across Architectures in Simulation:} In exhaustive simulation trials, our policy achieves $49.9\%$ success on V-Trellis and $46.0\%$ on UFO architectures, improving upon prior benchmark performance~\cite{JainRL}.
    \item \textbf{Real-World Sim-to-Real Transfer:} The learned policies successfully execute zero-shot across $28$ field trials and $10$ laboratory trials under variable outdoor environmental conditions.
    \item \textbf{RRT-Connect Baseline Comparison:} On physical hardware, our RL controller outperforms a classical RRT-Connect planner using reconstructed point clouds in overall success rate.
\end{itemize}
The following subsections detail these findings.

% This section presents the empirical findings from our evaluation of the pruning policy learned using Hybrid-PPO. We first detail the algorithmic validation in simulation, comparing our proposed method against established online and offline baselines to quantify the performance gains achieved by incorporating the offline dataset. Next, we present the results of our exhaustive policy evaluation across 3,000 simulated pruning generated using 100 unseen trees, establishing the success limits of the learned policy compared to the oracle RRT-Connect planner. Finally, we report the outcomes of our physical hardware deployments in both controlled laboratory settings and commercial orchard environments, showing zero-shot sim-to-real transfer and benchmarking of our approach against the classical motion planning baseline.

%% file: sections/9_1_algorithm_experiments.tex
\begin{table*}[ht]
\centering
\caption{ 95\% confidence intervals (CI) for success rates and differences in success rates with Hybrid-PPO using hierarchical bootstrapping. Significance is based on whether the CI for the difference includes 0.}
\begin{tabular}{lccc}
\hline
\textbf{Algorithm} & \textbf{Success Rate (95\% CI)} & \textbf{$\Delta$ Success Rate (vs. Hybrid-PPO) (95\% CI)} & \textbf{Statistically Significant?} \\
\hline
Hybrid-PPO & 0.477 [0.455, 0.500] & - & - \\
PPO        & 0.414 [0.353, 0.475] & 0.063 [-0.003, 0.118] & No \\
PPO+BC       & 0.425 [0.405, 0.446] & 0.051 [0.016, 0.088]  & Yes \\
\hline
\end{tabular}
\label{tab:bootstrapping}
\end{table*}
\input{figures/comparison_training_plots}

\subsection{Algorithmic Validation}  
\label{sec:results:algo}

Figure~\ref{fig:comparison_training} illustrates the training reward curves for each method across three random seeds. Standard PPO exhibits noticeably higher variance across seeds, whereas Hybrid-PPO and PPO+BC consistently achieve higher mean rewards, despite overlapping reward distributions across methods. 

Because raw reward magnitude does not always correlate directly with task execution---often due to misaligned reward shaping or potential reward hacking---we ground our primary evaluation on task success rates, which provide a direct measure of physical pruning performance.

Table~\ref{tab:bootstrapping} summarizes the task success rates alongside hierarchical bootstrap confidence intervals (CI) for all evaluated methods. Hybrid-PPO achieves the highest overall mean success rate ($47.7\%$), outperforming PPO+BC ($42.5\%$) and vanilla PPO ($41.4\%$).

Specifically, Hybrid-PPO yields a statistically significant mean improvement of $5.1\%$ over PPO+BC, demonstrating that our hybrid formulation provides tangible benefits beyond naive BC regularization. Against vanilla PPO, Hybrid-PPO achieves a mean improvement of $6.3\%$. While high seed-to-seed variance in vanilla PPO causes the $95\%$ paired bootstrap CI to slightly span zero, the distribution is centered substantially above zero, indicating consistent trend performance gains.

These results demonstrate that Hybrid-PPO effectively leverages offline demonstration data while outperforming standard RL baselines. Combining V-trace--style advantage correction with a clipped surrogate objective ensures stable updates even when off-policy dataset distributions diverge significantly from the target policy distribution.

%% file: figures/comparison_training_plots.tex
\begin{figure}[t]
    \centering
    \includegraphics[width=\columnwidth]{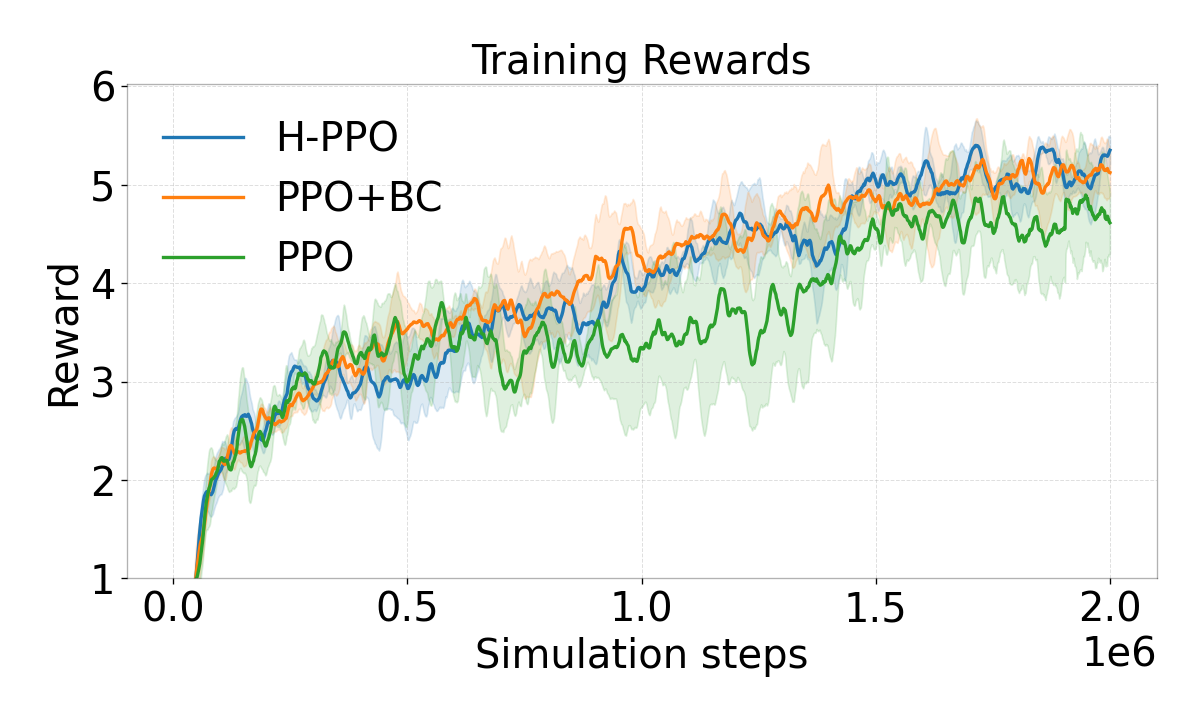}
   \caption{Training curves for Hybrid-PPO, PPO, and PPO+BC over three random seeds. The line represents the mean, and the shaded region indicates one standard deviation. Both PPO+BC and Hybrid-PPO achieve higher mean rewards by leveraging synthetic data, compared to online PPO.}
    \label{fig:comparison_training}
\end{figure}

%% file: sections/9_2_sim_experiments.tex
\input{figures/9_error_plots}
\subsection{Policy Evaluation across Architectures in Simulation}
\label{sec:results:policy}

Across $3,000$ evaluation trials in simulation, the learned policy achieved an overall success rate of $49.9\%$ on the V-Trellis architecture and $46.0\%$ on the UFO architecture. Notably, this  exceeds the $30\%$ success rate previously reported by Jain et al.~\cite{JainRL} on comparable V-Trellis structures. Figure~\ref{fig:violin_plots} (orange density plots) shows the distributions of reaching error (left), pointing error (middle), and perpendicularity error (right) for trained policies on V-Trellis (top) and UFO (bottom) canopies, with vertical green lines indicating the respective success thresholds.

Deconstructing individual termination criteria reveals that while the policy excels at spatial reaching, maintaining precise tool orientation poses a greater challenge. For V-Trellis trials, $96.9\%$ satisfied the reaching distance threshold, $54.0\%$ met the pointing alignment threshold, and $52.6\%$ satisfied the perpendicularity requirement, with $27.4\%$ of trials terminating due to environmental collisions. Similarly, for the UFO architecture, $96.7\%$ of trials met the reaching threshold, $52.9\%$ met pointing, and $49.8\%$ met perpendicularity, alongside a $24.0\%$ collision rate. These metrics demonstrate that achieving orthogonal alignment relative to the target branch is considerably more difficult for the policy than basic spatial reaching.

Serving as an approximate upper bound, an oracle RRT-Connect motion planner—operating with perfect ground-truth perception—achieved collision-free completion rates of $93.0\%$ on V-Trellis and $95.0\%$ on UFO architectures. This gap indicates that while our learned policy yields marked improvements over prior RL baselines, further gains in collision avoidance and orientation control are possible in dense clutter.

Visual inspection of failure cases indicates that while the policy effectively avoids obstacles within the end-effector camera's field of view, collisions frequently involve proximal manipulator links outside the camera frame. Mitigating these out-of-frame collisions in future work may require integrating exocentric global cameras or full-body tactile sensing. Further, scaling the synthetic dataset and larger simulation interactions have potential to improve this success rate and narrow the gap to oracle performance.
\input{figures/10_real_world_sankey}

%% file: figures/9_error_plots.tex
\begin{figure*}[t]
    \centering
    \includegraphics[width=\textwidth]{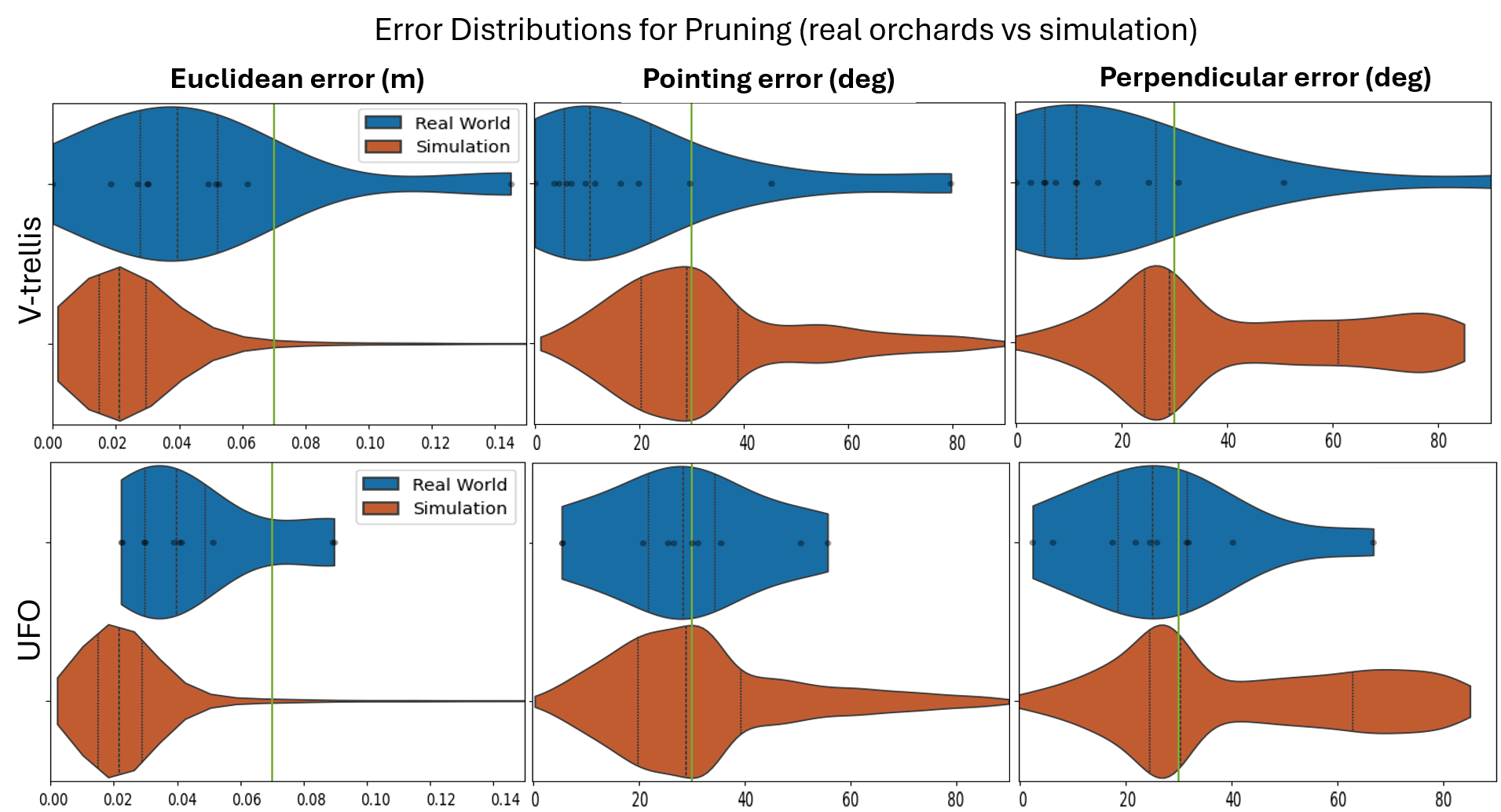}
    \caption{Error metrics for real‑world and simulated pruning trials on V-Trellis and UFO trees.
    \textit{(Top)} V-Trellis. \textit{(Bottom)} UFO. For each tree type, the upper violin in each panel shows real‑world results and the lower violin shows simulation results.
    From left to right, columns show (1) Euclidean error, (2) pointing error, and (3) perpendicularity error. The vertical green line in each panel denotes the success threshold for that metric and the individual samples for real-world trials are marked as black circles.}
    \label{fig:violin_plots}
\end{figure*}

%% file: figures/10_real_world_sankey.tex
\begin{figure*}[t]
    \centering
    \includegraphics[width=\textwidth]{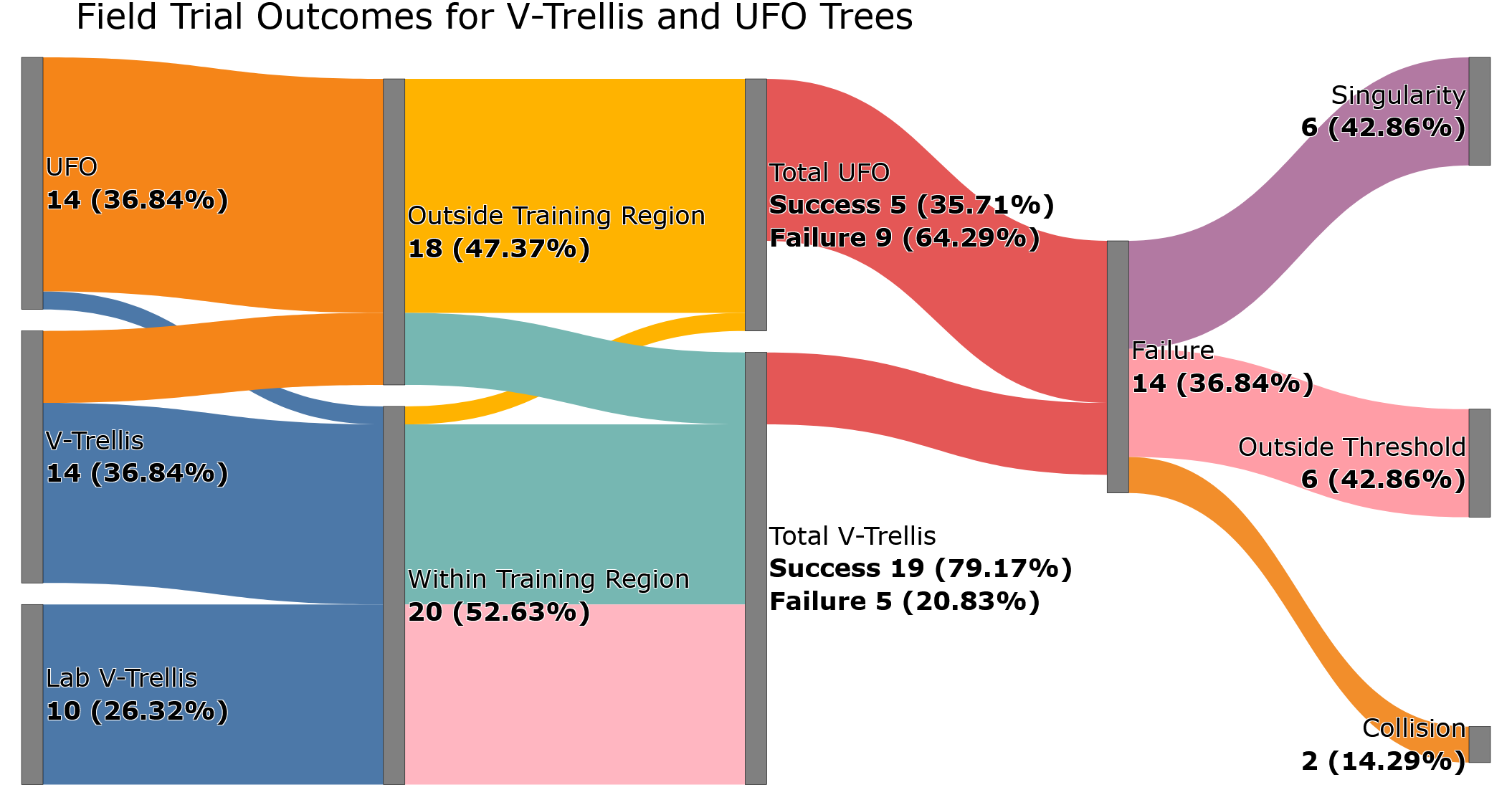}
    \caption{Sankey diagram showing the distribution of failure modes during real-world pruning trials. The first decision step categorizes trials based on whether the chosen cutpoint is within the specified task space. Subsequent failures are categorized as kinematic singularities (42.8\%), end-effector outside pruning thresholds (42.8\%), and collisions with tree structures (14.2\%). Percentages indicate the proportion of total failures attributed to each mode. 
}
    \label{fig:sankey}
\end{figure*}

%% file: sections/9_3_real_world_experiments.tex
\subsection{Real-World Sim-to-Real Transfer}
\label{sec:results:real_world}

Figure~\ref{fig:sankey} presents a detailed breakdown of success rates and failure modes observed during real-world physical deployments across both laboratory and operational orchard environments. We identified three primary failure modes during physical execution: (a) robot stoppage due to kinematic singularity limits, (b) timeout exceeding maximum allocation before satisfying pruning thresholds, and (c) hard collisions requiring human operator intervention. Detailed findings for all three deployment settings—the commercial V-Trellis orchard, the experimental UFO orchard, and the laboratory V-Trellis tree—are presented below.

\vspace{1ex}\noindent\textit{V-Trellis Apples in Commercial Orchard:} 
The Envy apple trees were trained on a V-Trellis architecture with the structural planar ``V" inclined at approximately $30^\circ$. Orchard geometry closely matched conditions modeled during simulation training, with most sampled target points ($10$ out of $14$) falling strictly within the learned task space volume (see Section~\ref{sec:meth:setting_up_learning}.\ref{sec:meth:setting_up_learning:sampling}). Under these conditions, the policy achieved a $71\%$ overall success rate ($10/14$ trials).

\vspace{1ex}\noindent\textit{UFO Cherries in Experimental Orchard:} 
The UFO cherry orchard comprised an experimental plot with trees trained at a steeper inclination of approximately $45^\circ$. This geometry introduced a significant structural domain shift relative to simulation training, violating geometric assumptions defining the primary task space. Consequently, nearly all target branches ($13$ out of $14$) lay outside the robot's pre-defined workspace volume. Despite this substantial spatial discrepancy, the policy exhibited encouraging zero-shot generalization, achieving a $35\%$ success rate ($5/14$ trials).

\vspace{1ex}\noindent\textit{V-Trellis Apples in Laboratory:} 
This setup was constructed using natural apple tree branches joined via custom 3D-printed connectors, allowing precise control over branch geometry while mounted to trellis wires spaced identically to commercial orchard standards (Fig.~\ref{fig:3d_reconstruction}). However, the tree was mounted vertically rather than on a $30^\circ$ incline, yielding a planar geometry that afforded greater kinematic reach. For these sim-to-real trials, target cutpoints were sampled within the reachable workspace and biased toward safer configurations matching training assumptions (Section~\ref{sec:meth:setting_up_learning}.\ref{sec:meth:setting_up_learning:sampling}). The policy achieved a $90\%$ success rate ($9/10$ trials). These $10$ laboratory trials are distinct from the separate $15$-point benchmark evaluation used for the RRT-Connect comparison below.

\vspace{1ex}\noindent\textbf{Execution Speed:} 
Across successful field and laboratory trials, the mean execution time to execute a complete pruning trajectory was $41.33 \pm 19.8\,\text{s}$. While slower than expert human manual pruning, this cycle time establishes a working baseline for fully autonomous robotic dormant pruning.

\vspace{1ex}\noindent\textbf{Failure Mode Analysis:} 
The blue density plots in Fig.~\ref{fig:violin_plots} depict error distributions from physical field trials in V-Trellis and UFO orchards. Because each distribution is based on a limited sample size (14 data points), we overlay the individual samples as semi-transparent markers. For the V-Trellis architecture, $92.8\%$ of trials met reaching distance thresholds, $85.7\%$ met pointing alignment thresholds, and $85.7\%$ met perpendicularity thresholds. For the UFO architecture, $78.5\%$ met reaching, $64.28\%$ met pointing, and $64.28\%$ met perpendicularity. These physical deployment results closely mirror trends observed in simulation: tool orientation alignment remains substantially more difficult to minimize than spatial end-effector positioning.

Analysis of failure modes (Fig.~\ref{fig:sankey}) indicates that singularity HALTs ($42.8\%$ of total failures) occurred because the physical UR5e arm controller triggers built-in joint velocity safety stops when the Jacobian condition number exceeds threshold limits. In contrast, the simulation policy was trained using a damped least-squares (DLS) IK controller, which allowed smooth local recovery through near-singular configurations without triggering hard stops. Timeout failures ($42.8\%$) correspond to episodes where the policy failed to converge within orientation or distance thresholds before reaching the maximum time allocation. Finally, collision failures ($14.2\%$) involved contact with rigid structural elements such as metal trellis wires, wooden posts, or main trunks and branches.

%% file: sections/9_4_real_world_baseline.tex
\subsection{RRT-Connect Baseline Comparison} 
\label{sec:results:real_world_baseline}

To rigorously evaluate the learned policy against a classical motion planning baseline, we conducted a direct comparison using the RRT-Connect planner guided by the 3D reconstructed collision mesh. Both methods were evaluated on an identical shared set of $15$ laboratory cutpoints sampled randomly for diversity in location and orientation, without the safety and reachability bias used in the $10$ sim-to-real laboratory trials above.

\vspace{1ex}\noindent\textbf{RL Policy Performance:} 
The learned policy achieved a $46.7\%$ success rate ($7$ out of $15$ trials). The $8$ failures consisted of $2$ unacceptable collisions, $4$ kinematic singularities, and $2$ instances where the end-effector remained outside the spatial thresholds at timeout. Notably, the majority of the singularity failures occurred when the policy attempted to reach deeply occluded branches within the inner canopy and at the lower trellis.

\vspace{1ex}\noindent\textbf{RRT-Connect Performance:} 
The RRT-Connect baseline achieved a $26.6\%$ success rate ($4$ out of $15$ trials). Out of the $11$ failures, $3$ were due to unacceptable collisions, $6$ were planning failures (planning time $> 60\,\text{s}$), and $2$ timed out outside the success spatial threshold. These failures were primarily driven by three perceptual limitations: (a) errors during point cloud alignment, where even minor spatial deviations resulted in out-of-threshold positioning or collisions, (b) adjacent branches merging during 3D reconstruction, creating inflated collision boundaries that blocked valid paths, and (c) the complete failure of the depth sensors to capture thin, occluded tertiary branches. The 3D reconstruction of a canopy section (Fig.~\ref{fig:3d_reconstruction}) highlights the extreme difficulty of resolving these fine structural details using standard depth sensors.

Our visuomotor policy outperforms the classical point-cloud-based planner on physical hardware. This performance gap is primarily driven by the fragility of the 3D reconstruction pipeline—which frequently fails to capture smaller tertiary branches and trellis wires, or erroneously merges closely spaced branches with similar visual textures—combined with the classical planner's inability to reliably navigate high-clutter environments.

The primary bottleneck for classical sampling-based planners lies in the fragility of explicit 3D reconstruction, where standard ICP pipelines frequently fail due to noisy depth estimation and the erroneous merging of adjacent branches. These depth and reconstruction limitations could potentially be overcome by leveraging modern depth predictors (e.g., FastFoundationStereo~\cite{fastfoundationstereo} and DepthAnythingV3~\cite{depthanything3}) and visual SLAM such as ORB-SLAM~\cite{orbslam} to generate higher-fidelity point clouds. Evaluating how these modern perception methods can enhance the robustness of classical motion planning for pruning in dense canopy environments at the cost of extra processing presents a promising direction for future research.
\input{figures/11_3d_reconstruction}

%% file: figures/11_3d_reconstruction.tex
\begin{figure*}
    \centering
    \includegraphics[]{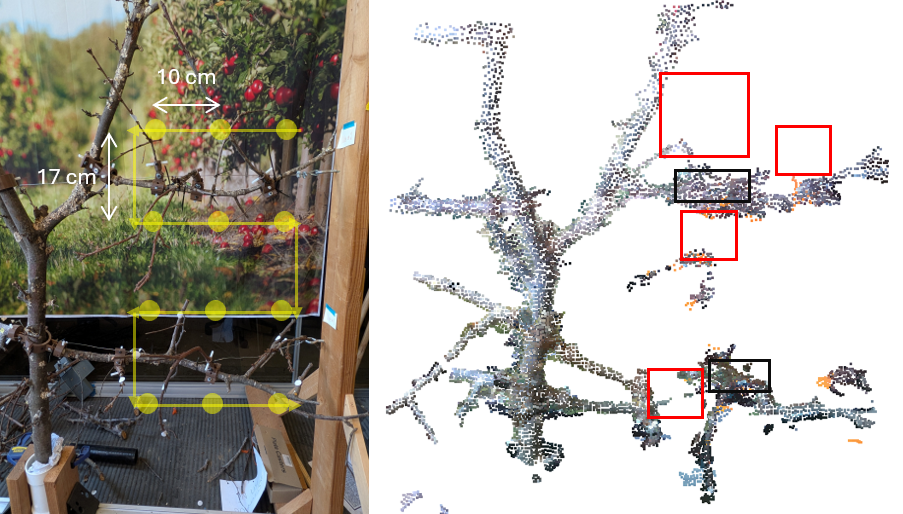}
    \caption{\textit{(Left)} Ladder scanning pattern employed to obtain views of the lab tree, the robot follows the arrows stops at the circle to obtain an RGB-D view and continues. \textit{(Right)}
    Colored reconstructed pointcloud built using the method described in Section~\ref{sec:experiments}.~\ref{sec:experiments:real_world_baseline}. The red boxes indicate missing branches while the black boxes indicate cluttered branches merging due to erroneous reconstruction.}
    \label{fig:3d_reconstruction}
\end{figure*}

%% file: sections/12_discussion.tex
\section{DISCUSSION AND FUTURE WORK}
\label{sec:limitations}

While our learned policy demonstrates improvement over classical planners in real-world deployments, its performance remains below that of the  oracle baseline. Analyzing these failure modes reveals several key limitations in our current approach, to be addressed by future work.

\vspace{1ex}\noindent\textit{Collision Avoidance:} 
Currently, the policy is not fully capable of robust collision avoidance relying solely on eye-in-hand camera observations. In camera-frame collisions are avoided through an exclusively egocentric view. However, the proximal links of the robot are left out of perception thus vulnerable to collisions. Achieving deployment-ready reliability will require integration of additional sensory modalities, such as an exocentric global camera, or force/torque feedback to provide a holistic representation of the manipulator within the cluttered canopy. Further, such learning-based methods can be employed along with classical collision avoidance methods~\cite{marcustemplating} or for final alignment~\cite{lukealignment} for robustness. Further, use of modern depth foundation models~\cite{fastfoundationstereo, depthanything3} may provide better perception of structure as compared to Optical Flow which inherently relies on motion.

\vspace{1ex}\noindent\textit{Kinematic Limits:} 
The physical constraints of the 6-DOF UR5e manipulator introduce challenges in both reachability and achieving the large end-effector poses required for pruning. In orchard environments the arm frequently encounters kinematic singularities when reaching branches farther away. This severely restricts feasible motion trajectories and reduces the effective workspace the robot can operate in, largely reduced to the outer canopy of the tree. We require arms designed specifically keeping the planar tree structure in mind for better control. 

\vspace{1ex}\noindent\textit{Sim-to-Real Compliance Gap:} 
Our current simulation pipeline treats all tree structures as strictly rigid bodies. In actual agricultural environments, branches are inherently compliant and can often be safely pushed aside by the manipulator to reach occluded target cutpoints. Modeling this physical compliance in simulation could close a significant sim-to-real gap, allowing the policy to learn efficient reaching behaviors.

\vspace{1ex}\noindent\textit{Algorithmic and Computational Scaling:} 
Hardware compute constraints inherent to a CPU-based simulator restricted our training to 2 million online steps---a relatively small budget compared to state-of-the-art RL tasks that typically leverage billions of interactions. Transitioning to a highly parallelized, GPU-accelerated simulation environment, such as Isaac Lab~\cite{isaac_lab} or MuJoCo~\cite{mujoco_mjx}, would alleviate this bottleneck and allow for massive scaling of simulator interactions. Additionally, algorithmic augmentations to the Hybrid-PPO framework, such as incorporating the prioritized replay buffers commonly utilized in hybrid off-policy algorithms~\cite{vecerik2017leveraging, hester2018deep}, offer a promising avenue for non-trivial performance gains.

%% file: sections/13_conclusion.tex
\section{CONCLUSION}
\label{sec:conclusion}

We presented a comprehensive framework for learning closed-loop, optical-flow-based visuomotor policies for autonomous dormant pruning in modern planar orchards. Our pipeline integrates a procedural synthetic tree generator, a physics-based orchard simulator, automated motion-planning demonstration generation, and a Hybrid-PPO algorithm that seamlessly combines offline expert trajectories with online rollouts. We demonstrated that policies trained entirely in simulation achieve successful zero-shot sim-to-real transfer to a physical robotic arm operating in commercial and experimental orchards across two distinct canopy architectures (V-Trellis and UFO). Autonomous dormant pruning remains an exceptionally challenging task due to severe visual occlusion, dense physical clutter, and strict end-effector orientation requirements.  Despite these challenges, our learned policy achieves an absolute success rate of approximately 50\%---roughly half the performance of an idealized oracle planner equipped with perfect state information. These results validate the viability of data-driven reinforcement learning for complex, clutter-intense agricultural manipulation.